\documentclass[letterpaper,10 pt,conference]{ieeeconf}  % Comment this line out if you need a4paper
\IEEEoverridecommandlockouts                              % This command is only needed if
\usepackage{cite,color,algorithm,algorithmic}
\usepackage{graphics,multirow} % for pdf, bitmapped graphics files
\usepackage{graphicx,subfigure,epstopdf}
\usepackage{epsfig} % for postscript graphics files
\usepackage{mathptmx} % assumes new font selection scheme installed
\usepackage{times} % assumes new font selection scheme installed
\usepackage{amsmath} % assumes amsmath package installed
\usepackage{amssymb}  % assumes amsmath package installed
\usepackage{hyperref}

\usepackage{cleveref}
\crefname{section}{Sec.}{Sec.}
\crefname{figure}{Fig.}{Fig.}
\crefname{table}{TABLE}{TABLE}
\crefname{algorithm}{\textbf{Algorithm}}{\textbf{Algorithm}}
\crefname{equation}{Eq.}{Eq.}

\usepackage{siunitx}

\usepackage[misc]{ifsym}
\hypersetup{hidelinks,
	colorlinks=true,
	allcolors=blue,
	pdfstartview=Fit,
	breaklinks=true}

\title{\LARGE \bf
Adaptive Environment Modeling Based Reinforcement Learning for Collision Avoidance in Complex Scenes
}

\author{Shuaijun Wang$^{1,2,*}$, Rui Gao$^{2,*}$,  Ruihua Han$^{2,3}$, Shengduo Chen$^{2}$, Chengyang Li$^{2}$ and Qi Hao$^{2,4,\dag}$% <-this % stops a space
\thanks{This work is partially supported by the Science and Technology Innovation Committee of Shenzhen City (No: JCYJ20200109141622964).}
\thanks{$^{*}$ indicates equal contributions.}
\thanks{$^{\dag}$ Corresponding author: Qi Hao (hao.q@sustech.edu.cn).}
\thanks{$^{1}$Harbin Institute of Technology, Harbin, Heilongjiang, China, 150001.}%
\thanks{$^{2}$Department of Computer Science and Engineering,
Southern University of Science and Technology, Shenzhen, Guangdong, China, 518055.}
\thanks{$^{3}$Department of Computer Science, The University of Hongkong, 999077.}%
\thanks{$^{4}$Rsearch Institute of Trustworthy Autonomous Systems,
 Southern University of Science and Technology, Shenzhen, Guangdong, China, 518055.}%
}

\begin{document}
\maketitle
\thispagestyle{empty}
\pagestyle{empty}
%%%%%%%%%%%%%%%%%%%%%%%%%%%%%%%%%%%%%%%%%%%%%%%%%%%%%%%%%%%%%%%%%%%%%%%%%%%%%%%%%
\begin{abstract}
The major challenges of collision avoidance for robot navigation in crowded scenes lie in accurate environment modeling,
fast perceptions, and trustworthy motion planning policies. This paper presents a novel adaptive environment model based
collision avoidance reinforcement learning (i.e., AEMCARL) framework for an unmanned robot to achieve collision-free motions
in challenging navigation scenarios. The novelty of this work is threefold: (1) developing a hierarchical network of gated-recurrent-unit (GRU) for environment modeling; (2) developing an adaptive perception mechanism with an attention module; (3)
developing an adaptive reward function for the reinforcement learning (RL) framework to jointly train the environment model, perception function and motion planning policy.
The proposed method is tested with the Gym-Gazebo simulator and a group of robots (Husky and Turtlebot) under various crowded scenes. Both simulation and experimental results have demonstrated the superior performance of the proposed method over baseline methods.
\end{abstract}
%
%
%%%%%%%%%%%%%%%%%%%%%%%%%%%%%%%%%%%%%%%%%%%%%%%%%%%%%%%%%%%%%%%%%%%%%%%%%%%%%%%%%
\section{INTRODUCTION}
\begin{figure}[t]
 \centering
 %\resizebox{\linewidth}{!}{
 \includegraphics[width=0.4\textwidth]{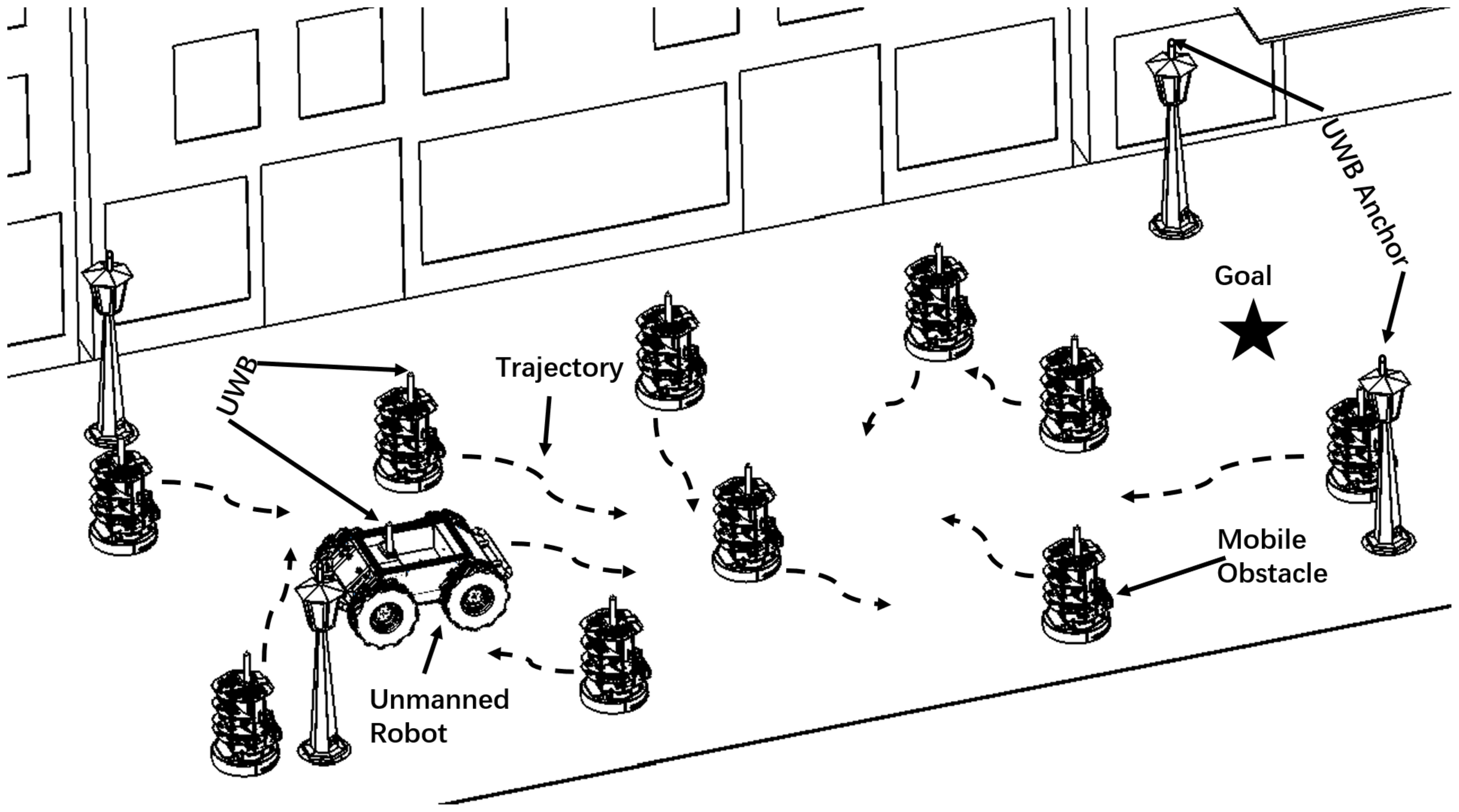}\\ \vspace{-0.3cm}
  \caption{An illustration of an unmanned robot with the capability of collision avoidance in crowded scenes (of robotic mobile obstacles). All robots are equipped with Ultra-wideband (UWB) modules for distance measurement and localization.}
     \label{fig:poster}
     \vspace{-0.5cm}
\end{figure}

End-to-end RL-based unmanned mobile robots are advantageous in the high degree of autonomy, robustness against unmodeled uncertainties, and high-speed decision~\cite{SADRL,ma2021reinforcement,han2022reinforcement}.
The main components of RL-based unmanned robots include (1) sensing, motion and communication units, (2) world and robot models, (3) policy and probability based reward functions, and (4) state and action spaces.
Developing RL-based autonomous robot systems for crowded scenes, as shown in \cref{fig:poster}, has to deal with the following technical challenges:
\begin{enumerate}
  \item \textbf{Dynamic environment modeling}. In the real world, many objects such as pedestrians, vehicles, animals move around with various behaviors and interactions; how to develop a hierarchical model to represent such a dynamic environment in different degrees of complexity is still a challenging problem;
 % For a high-speed robot, motion decisions have to be made as soon as possible; how to reduce the computational complexity of the RL neural network becomes critical;
  \item \textbf{Adaptive perception mechanism}. The robot should be able to perceive the environment with an attention mechanism, and adaptively use computational resources according to perception confidences.%within a certain degree of perception confidences.
  \item \textbf{Collision-free policy for motion planning}. The quality of RL-based motion planning policies relies on the selection of reward functions and state representation. An ideal reward function should reduce not only the collision possibility between the robot and the nearest obstacle but also that between the robot and neighboring obstacles within a time window.
\end{enumerate}

Most RL-based crowd-aware approaches use simple dynamic environment models such as multi-layer perceptron (MLP), which cannot fully describe real-world complexities and uncertainties. Meanwhile, perception functions have to trade off between decision speed and environment modeling accuracy by selecting proper structures of deep learning models~\cite{chen2020relational}. On the other hand, reward function and attention mechanisms~\cite{chen2019crowd,chen2020relational} have been developed to make better decisions with reduced computational workloads. However, no such a RL-based framework has been developed yet, which contains high-order environment models as well as adaptive perception, adaptive reward function, and attention modules.

\begin{table*}[htbp]
  \centering
  \caption{A brief summary of Traditional and Reinforcement learning based collision-free navigation methods}
\resizebox{0.9\linewidth}{!}{
  \begin{tabular}{c|c||c|c|c||c|c|c||c}
  \hline \hline
  \multirow{3}{*} {Scheme} & \multirow{3}{1.5cm} {\centering Navigation Method}& \multicolumn{3}{c||}{$\textbf{Dynamic Environment Modeling}$} & \multicolumn{3}{c||}{$\textbf{Adaptive Perception Mechanism}$} &\multicolumn{1}{c}{$\textbf{Collision-free Policy}$}             \\
     \cline{3-9}
  ~ & ~ & \multirow{2}{1cm} {\centering Agent number} & \multirow{2}{1.5cm} {\centering Multi agent representation} & \multirow{2}{1.5cm} {\centering Hierarchical representation} & \multirow{2}{1.5cm} {\centering Perception confidence} &
  \multirow{2}{1.5cm} {\centering Self-tuning} & \multirow{2}{1.5cm} {\centering Attention mechanism} %&\multirow{2}{2cm} {\centering CA (Learning) scheme}
  &  \multirow{2}{*} {Reward}  \\
  ~ & ~ & ~ & ~ & ~ & ~ & ~ & ~ & ~  \\ \hline
   \multirow{2}{2cm}{\centering \textbf{Traditional Approach}}
    & RVO~\cite{RVO2008}    & Arbitrary & Math modeling & $\times$ & $\times$ &  $\times$ & $\times$  & N/A \\
     \cline{2-9}
  ~ & ORCA~\cite{ORCA}      & Arbitrary & Math modeling & $\times$ & $\times$ & $\times$ & $\times$  & N/A  \\ \hline
    \multirow{5}{2cm}{\centering \textbf{RL-based Approach}}
    &LSTM-RL~\cite{everett2018motion}& Fixed Number & LSTM \& MLP & $\times$ & $\times$ & $\times$ & $\times$  & +   \\
%         \cline{2-9}
%   ~  & LSTM-RL~\cite{everett2018motion}& High & LSTM \& MLP & $\times$ & $\times$ & $\times$ & $\times$  & +   \\
        \cline{2-9}
  ~ & CADRL~\cite{CADRL2017} $\setminus$ SARL~\cite{chen2019crowd}        & Fixed Number$\setminus$High  & MLP         & $\times$ & $\times$ & $\times$ & $\times \setminus \surd$  & +    \\
      \cline{2-9}
%   ~ &   & High & MLP         & $\times$ & $\times$ & $\times$ & $\surd$   & +    \\
%   \cline{2-9}
  ~ & RGL~\cite{chen2020relational}$\setminus$G-GCNRL~\cite{liuming2020robot}& Fixed Number$\setminus$High & GNN$\setminus$GCN      & $\times$ & $\times$ & $\times$ & $\surd$  & ++$\setminus$+   \\
   \cline{2-9}
  ~ & SODARL~\cite{liu2020DRL}& Arbitrary & CNN       & $\times$ & $\times$ & $\times$ & $\surd$  & ++   \\
      \cline{2-9}
  ~ & AEMCARL (Ours)               & Arbitrary & GRU        & $\surd$  & $\surd$ & $\surd$  & $\surd$   & ++     \\ \hline \hline
  \end{tabular}
  }
  \begin{tabular}{p{17cm}}
  The number of + reflects the performance of this functionality of the method.
    Symbol $\surd$: This functionality is supported. Symbol $\times$: This functionality is not supported or not considered.
    CA: Collision avoidance. GNN: Graph neural network. GCN: Graph convolution network. FC: Fully connection. The angular map in SODARL plays a same role with the attention mechanism.
  \end{tabular}
  \label{Tab:briefIntrouction}
  \vspace{-0.7cm}
  \end{table*}

%In addition, the prediction capability and an attention mechanism of the RL framework are essential and challenging factors for a robust collision-free policy. And an attention mechanism of the RL framework can contribute to boosting the speed of decision making. Integrating predictive modeling with an attention mechanism into the RL framework is also a challenge, which refers to the computation complexity being exponential proportional to the number of the agents.
%The model with prediction capability plays an important role in understanding the physical, and crowded world, which can promote the collision avoidance performance of the robot especially in the perception sensors failed.

In this paper, we propose a robust RL-based motion planning framework with an adaptive environmental model (AEM) and adaptive reward function to achieve collision-free navigation in crowded scenarios.
The estimated perception confidences are used to change the AEM structure in real-time.
Adaptive reward function and attention modules are developed to improve the generalization capability of the system. The main contributions of this paper include
%  percept the global information by integerating pairwise local interaction information.
% explore the deeper and potential information between both robot-agent and agent-agent. Moreover, the global structural global information is obtained by the local pairwise based information and used to decide the order of motion policy.
\begin{enumerate}
\item {We design} a novel RL-based motion planning framework, which consists of modules of modeling, attention, and action, to achieve real-time navigation with collision avoidance;
\item {We propose} a hierarchical environment model to represent multiple agents of dynamic environments;
\item {We develop} a perception confidence based adaptation mechanism as well as an attention module to achieve fast decision with reduced computational complexity;
\item {We propose} an adaptive reward function reward function and a set of robust navigation policy training algorithms. The open-source code is available at:\\
\href{https://github.com/SJWang2015/AEMCARL}{https://github.com/SJWang2015/AEMCARL}.
 %Real-time implementation of the proposed framework on a Husky robot.
\end{enumerate}

The rest of this paper is organized as follows. Section~\ref{Realtedwork} introduces the related work on RL-based motion planning.
Section~\ref{setup-statement} describes the system setup and problem statement. Section~\ref{Approach} presents the proposed method.
Section~\ref{experiments} provides the experiment results and ablation studies. Section~\ref{conclusions} concludes this paper and outlines future work.
\section{RELATED WORK}
\label{Realtedwork}
\cref{Tab:briefIntrouction} summarizes a number of navigation methods with collision avoidance for a single robot in crowded scenes.
Compared with conventional methods, the RL-based approaches can learn from the various simulation experiences to generate robust and high-efficiency motion planning policies.
The most challenging problem for motion planning in crowded scenes is to build dynamic environment models (DEMs) which should accommodate a large number of mobile robots, and describe the spatial and temporal interactions among agents.
Most DEMs can be classified into two groups: object-centric and relation-centric.
The former emphasizes the pose and velocity information of agents~\cite{battaglia2016IN}; while the latter describes the current and future relationships among agents~\cite{battaglia2016IN}.
To achieve better modeling performance, fully connected (FC) layers and  long short-term memory (LSTM)~\cite{LSTM1997} units have been used to describe long-term and long-range relation-centric interactions among agents~\cite{han2022reinforcement,li2021rain}.
Compared with LSTM and GRUs~\cite{gru2014}, Transformer ~\cite{vaswani2017attention} don't rely on the order of input data, and can capture the relations between participants precisely. The graph networks have also been used to build the relationships of all agents in the environment~\cite{chen2020relational,liuming2020robot}, which are flexible to combine with the attention mechanism, and help the network achieve better obstacle avoidance performance. To inference the dynamic environmental relationship in the future, the graph must be fixed, or the drastically changing number of obstacles might cause the algorithm's performance to be unstable.
% gated recurrent units (GRU)~\cite{gru2014} are advantageous in high computational efficiency and fast convergence. However, no hierarchical GRU or extended GRU (EGRU) networks have been developed to represent complex dynamic environments with a large number of agents.

Recurrent and feedforward networks~\cite{chen2019crowd,CADRL2017} with high order structures have been used to filter and model the complex dynamic environments. However, in a typical crowded scene, the number of mobile obstacles which might cause collisions might vary from time to time. The prediction of the interactions among a large number of agents usually incur high computational workloads. Attention mechanisms can help the unmanned robot to focus on the most threatening mobile and static obstacles in the environment, which are not necessarily correlated to the robot-obstacle distance~\cite{vemula2018social, liu2020DRL,sathyamoorthy2020densecavoid}. On the other hand,
it is necessary to develop a set of self-tuning mechanisms, in which the environment model structure can be changed according to the perception confidences~\cite{ACT2016}. Despite many efforts in this aspect having been used for natural language processing~\cite{ACT2016,vaswani2017attention}, little work has been done to achieve online self-tuning DEMs for collision avoidance.

Compared with policy-based and actor-critic RL approaches such as PPO~\cite{PPO} and A3C~\cite{A3C2016}, the value-based off-policy Deep Q-learning network (DQN) are advantageous in simple structure, low computational complexity, and fast training convergence. However, it is very critical to choose proper reward functions for DQN based collision-free navigation policies, which should avoid being stuck in freezing points~\cite{sathyamoorthy2020frozone} and move forward to the goal position as fast as possible. Besides, proper representations of the interactions among critical agents under attention are also helpful for network training convergence~\cite{chen2019crowd, sathyamoorthy2020densecavoid}. Inspired by using anticipatory behaviors to increase the robustness of the collision-free policy~\cite{sathyamoorthy2020densecavoid}, we design an adaptive reward function based on predictive collision probabilities using estimated velocities of neighboring obstacles within a fixed time window.

In this work, a value-based RL method, AEMCARL, is proposed to achieve collision-free navigation policies. AEMCARL uses a hierarchical environment model (HEM) with an {adaptive perception mechanism (APM) and an attention module} to learn the relationships among multiple agents. An adaptive reward function and an attention module are also used to improve the computational efficiency and training stability.
\section{System Setup and Problem Statement}
\label{setup-statement}
\subsection{System Setup}
\begin{figure*}[htbp]
  \centering
  \resizebox{0.85\linewidth}{!}{\includegraphics[height=0.5\textwidth]{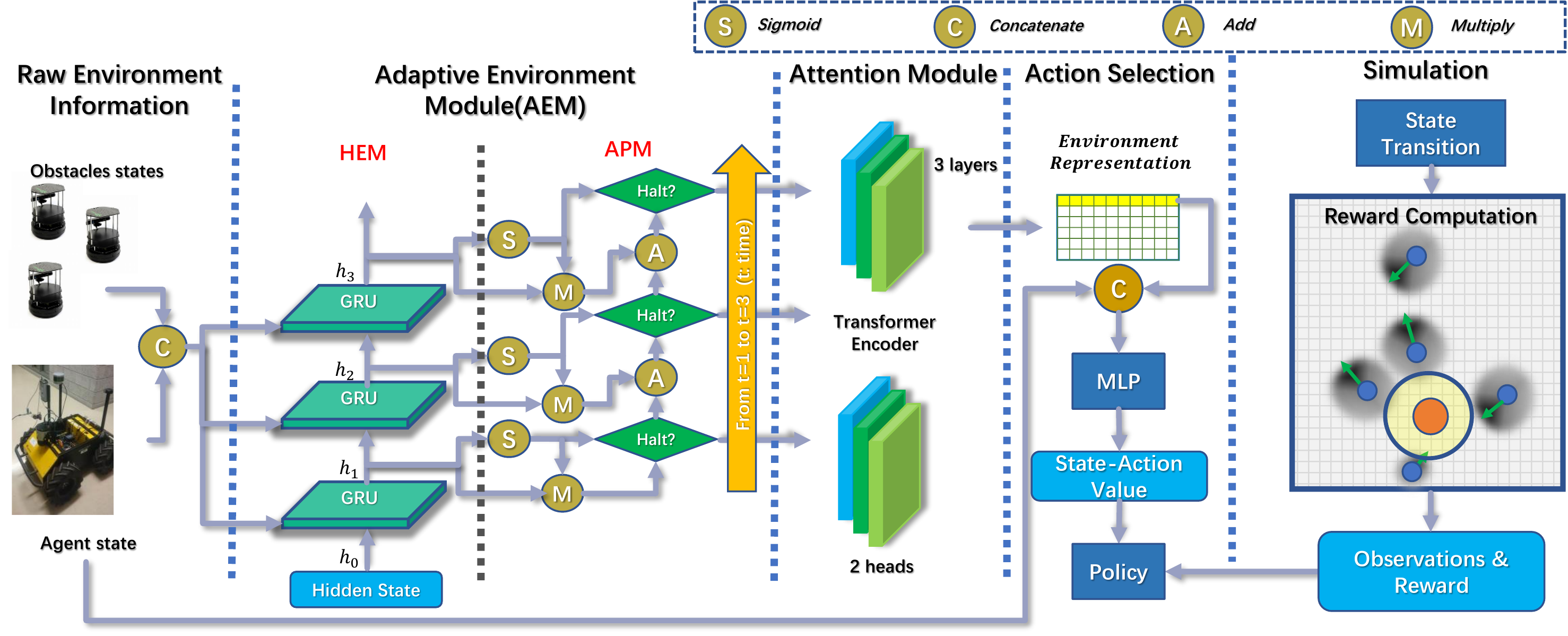}}\\  \vspace{-0.3cm}
  \caption{The system diagram of the proposed adaptive environment model RL (AEMCARL) based crowded-aware collision-free motion planning network. The first row with yellow color of the transformer encoder output means the global environment perception information of the robot, the other rows with white color correspond to the perception information of obstacles.}
      \label{fig:rl-model}
      \vspace{-0.5cm}
 \end{figure*}
\cref{fig:rl-model} illustrates the proposed RL-based navigation system with collision avoidance in crowded scenes for an
unmanned robot. The experiment platform contains a Husky as the unmanned robot and a group of Turtlebots as mobile obstacles. All robots are equipped with UWB modules, along with stationary UWB anchors, for localization, as shown in \cref{fig:poster}. All the odometry readings, including poses and velocities, UWB readings, and action commands are exchanged through the ROS communication network. The Gym-based simulator~\cite{chen2019crowd} is
used to train the hierarchical environment model (HEM), adaptive perception mechanism (APM), policy function, and evaluation network.
\subsection{Problem Statement}
\label{ProblemStatement}
%The state of the agent at time $t$ is denoted as $\textit{\textbf{$s_t$}}$, which includes velocity, $(\textit{$v_x^a$},\textit{$v_y^a$})$, position, $(\textit{$x^a$}, \textit{$y^a$})$, goal position, $(\textit{$g_x^a$},\textit{$g_y^a$})$, and preferred velocity value, $\textit{$v^{pref}_r$}$. The states of mobile obstacles at time $t$ is denoted as $\textit{\textbf{$o_t$}}$, \textcolor{red}{which contains all obstacles' states} include the velocity, $(\textit{$v_x^o$},\textit{$v_y^o$})$, position, $(\textit{$x^o$}, \textit{$y^o$})$, and preferred velocity value, $\textit{$v^{pref}_o$}$. \textcolor{red}{The shape of $\textit{\textbf{$o_t$}}$ is Nx5, N is the count of obstacles}. The input of the RL network is the joint state of the agent and \textcolor{red}{environment representation which is the output of AEM and attention module} \sout{obstacles} at time $t$, named ${env_t}$ , $\textit{\textbf{$s_t$}}=\textbf{(\textit{$e_t$}, \textit{$env_t$})}$, is defined by Eq.~(\ref{Eq:stat}), \textcolor[rgb]{1,0,0,}{and the output of the RL network is an action $\mathbf{a}$ composed by $(\textit{$v_x$},\textit{$v_y$})$.}

%The states of mobile obstacles at time $t$ is denoted as $\textit{\textbf{$o_t$}}$, \textcolor{red}{which contains all obstacles' states} include the velocity, $(\textit{$v_x^o$},\textit{$v_y^o$})$, position, $(\textit{$x^o$}, \textit{$y^o$})$, and preferred velocity value, $\textit{$v^{pref}_o$}$. \textcolor{red}{The shape of $\textit{\textbf{$o_t$}}$ is Nx5, N is the count of obstacles}.

The state of the agent at time $t$ is denoted as $\textit{\textbf{$ego_t$}}$ defined by \cref{Eq:stat}, which includes velocity, $(\textit{$v_x^a$},~\textit{$v_y^a$})$, position, $(\textit{$x^a$},~\textit{$y^a$})$, goal position, $(\textit{$g_x^a$},~\textit{$g_y^a$})$, and preferred velocity, $\textit{$v^{pref}_r$}$.
The states of mobile obstacles at time $t$ is denoted as $\textit{\textbf{$o_t$}} \in \mathbb{R}^{N*D_{obs}}$ ($N$ is the number of obstacles, $D_{obs}$ is feature size of obstacle), which contains states of all obstacles include the velocity, $(\textit{$v_x^o$},~\textit{$v_y^o$})$, position, $(\textit{$x^o$},~\textit{$y^o$})$, and preferred velocity value, $\textit{$v^{pref}_o$}$ defined by~\cref{Eq:stat}.
%and it's shape is N * $D_{obs}$ (N is the number of obstacles, $D_{obs}$ is feature size of obstacle).
The input of the RL network is the joint state of the agent and all obstacles, $\mathbf{s_t}=\textit{($ego_t$,$~o_t$)} \in \mathbb{R}^{(N+1)\times \left(D_{obs}+D_{agent}\right)}$. $\mathbf{s_t}$ contains the status of $N$ obstacles and an agent.
The output of the RL network is an action noted as $\mathbf{a}$ composed by $\textit{$v_x$}$ and $~\textit{$v_y$}$.
\begin{equation}\label{Eq:stat}
    \begin{aligned}
    ego_t &= \left[v_x^a, v_y^a, x^a, y^a, g_x^a, g_y^a, v^{pref}_a\right], \\
    o_t &= \left[v_x^o, v_y^o, x^o, y^o, v^{pref}_o\right]. \\
    \end{aligned}
\end{equation}
% \begin{equation}\label{Eq:stat}
% % \begin{array}{c}
% ego_t = \left[v_x^a, v_y^a, x^a, y^a, g_x^a, g_y^a, v^{pref}_a\right], \\
% %env_t = \left[v_x^o, v_y^o, x^o, y^o, v^{pref}_o\right]
% %\sout{o_t = \left[v_x^o, v_y^o, x^o, y^o, v^{pref}_o\right].}
% % \end{array}
% \end{equation}

% \begin{equation}\label{Eq:obstacles_stat}
% % \begin{array}{c}
% o_t = \left[v_x^o, v_y^o, x^o, y^o, v^{pref}_o\right]. \\
% %env_t = \left[v_x^o, v_y^o, x^o, y^o, v^{pref}_o\right]
% %\sout{o_t = \left[v_x^o, v_y^o, x^o, y^o, v^{pref}_o\right].}
% % \end{array}
% \end{equation}

The objective of the RL learning is to find the optimal motion planning policy, $Q^*_{\pi}\left(\textit{$\textbf{$s_t$}$},\textit{\textbf{$a_t$}}\right) := max~Q_{\pi}\left(\textit{$\textbf{$s_t$}$},\textit{\textbf{$a_t$}}\right)$, as shown in \cref{Eq:optimalreward}~\cite{chen2019crowd}, which can enable the agent to reach the destination as soon as possible with low collision probability in crowded scenes.
\begin{equation}\label{Eq:optimalreward}
\begin{aligned} Q^*_{\pi}\left(\textit{$\textbf{$s_t$}$},\textit{\textbf{$a_t$}}\right)=& \underset{\mathbf{a}_{t}}{\operatorname{argmax}} R\left(\mathbf{s}_{t}, \mathbf{a}_{t}\right)+\\ & \gamma^{\Delta t \cdot v^{pref}} \int_{\mathbf{s}_{t+\Delta t}^{j n}} P\left(\mathbf{s}_{t}, \mathbf{a}_{t}, \mathbf{s}_{t+\Delta t}\right) V^{*}\left(\mathbf{s}_{t+\Delta t}\right) d \mathbf{s}_{t+\Delta t}, \end{aligned}
\end{equation}
where $a_t$ is the action taken by the robot at the time $t$, $\gamma$ is the discount factor, $\gamma \in [0, 1]$, and $V^*$ is the optimal state-action value obtained by the trained value network, $P\left(\mathbf{s}_{t}, \mathbf{a}_{t}, \mathbf{s}_{t+\Delta t}\right)$ is the probability of transiting to $\mathbf{s}_{t+\Delta t}$ from $\mathbf{s}_{t}$ and $\mathbf{a}_{t}$.

Specifically, this work focuses on solving the following problems:
%1) how to develop a model to represent the states of agents in a dynamic environment; 2) how to reduce computational complexity of the RL neural networks with the multi-agent dynamic environment models; 3) how to achieve the collision free robust policy by adaptively selecting the number of layers of deep learning models. It can be concluded as the following formulas:
\begin{enumerate}
  \item How to develop a model to represent the dynamic interactions among multiple agents with high accuracy and efficiency;
  \item How to reduce the computational complexity of the system through the adaptive perception and the attention modules;
  \item How to achieve the collision-free robust motion planning policy using a proper reward function.
\end{enumerate}
%\vspace{-2.0cm}
%\begin{equation}\label{Eq:prbstat}
%\begin{array}{c}
%\pi^{*}\left(\mathbf{s}_{t}\right), where ~y_{t}^{n} = \frac{1}{N(t)}\sum_{n=1}^{N(t)} p_{t}^{n}h_{t}^{n}, \\
%\emph{s.t.}~ N(t) = \min\left\{n^{'}:\sum_{n=1}^{n^{`}} p_{t}^{n}\geq 1-\epsilon\right\},
%\end{array}
%\end{equation}
%where $p_{t}^{n}=\sigma \left( W^{h}h^{n}_{t}+b_{h} \right)$, $\sigma(\cdot)$ is the sigmoidal halting unit, $y_{t}^{n}$ is the output of the AEM module, $s_{t}^{n}$ is the input of the $n^{th}$ AEM module, $h^{n}_{t}$ is an hidden state, $N(t)$ is the number of computation step, and $\epsilon$ is the halt parameter.
%\begin{eqnarray}\label{Eq:prbstat}
%\pi^{*}\left(\mathbf{s}_{t}\right), where s_{t} = \sum_{n=1}^{N(t)} p_{t}^{n}y_{t}^{n},\\
%\emph{s.t.} ~
%N(t) = \min\left\{n^{'}:\sum_{n=1}^{n^{`}} h_{t}^{n}\geq 1-\epsilon\right\}
%\end{eqnarray}

% p_{t}^{n}=\sigma \left{ W^{h}h^{n}_{t}+b_{h} \right} p_{t}^{n} =\sigma \left(W^{h}h^{n}_{t}+b_{h}\right)
\section{Proposed Methods}
\label{Approach}
%The model-free RL methods can get a good performance and strong extendability in the new environment. Therefore, this work adopts the model-free RL method to implement the unmanned robot automatically pilot without collision in the real world.
%The RL method has the capability of smart exploration for the future unknown world to help the agent make more better policy and deeply exploitation for the previous experiences to help the agent chooses the action with high reward.
\subsection{Hierarchical Environment Model}
The action instantly taken by the agent is affected by both direct robot-to-obstacle interaction and potential obstacle-to-obstacle interaction. The goal of each obstacle in the real world is unpredictable and random.
%Therefore, each agent may be affected by a different number of agents when making a motion decision.
Inspired by the method~\cite{ACT2016}, the proposed HEM consists of a number of GRUs to represent the dynamic environment of multiple mobile obstacles, as shown in \cref{fig:rl-model}. Each GRU~\cite{gru2014} is given by

\begin{equation}
\label{gru}
\begin{aligned}
    z_t &= \sigma \left({\rm MLP}([h_{t-1}, s_t], W_z) \right),\\
    r_t &= \sigma \left({\rm MLP}([h_{t-1}, s_t], W_r) \right),\\
    q_t &= \tanh \left({\rm MLP}([r_t \odot h_{t-1}, s_t], W_q) \right),\\
    h_t &= \left(1 - z_t\right) \odot h_{t-1} + z_t \odot q_t,
\end{aligned}
\end{equation}
where $s_t$ is the input state defined in \cref{Eq:stat}, each MLP consists of two linear perception layers, and all these MLPs have the same structure. The output of each GRU is given by
\begin{equation}\label{eq:hidden}
h_t^n = \left\{
  \begin{array}{ll}
{\rm GRU}(h_0, s_t^1), & \textnormal{if } n=1  \\
{\rm GRU}(h_{t}^{n-1}, s_t^n), & \textnormal{otherwise}
  \end{array}
  \right.,
\end{equation}
where {$h_0$} is initialized by a zero vector,  $n$ is the number of computation iterations of the HEM, and {$h_t$} is the $t^{th}$ environment input of the HEM module.
The HEM has three layers, and each includes a GRU. The execution order of these GRUs is serial, and each GRU takes the output of the previous GRU and raw environment state as input. The output of GRU will be used to determine whether the output is enough to represent the raw environment state or not, if enough, the serial execution will be terminated, otherwise, continue the serial execution. The details of how to decide if enough are introduced in \cref{sec:APM}.

Since the linear transformation operation, as shown in \cref{eq:linear_mlp}, is used as the MLP operation of GRU, the feature size of input shape, denoted by $\left(D_{obs}+D_{agent}\right)$, must be equal to the number of rows of the learnable weights, denoted by $\bf{W}$, which can make each GRU module to take the different number of obstacles.
\begin{equation}\label{eq:linear_mlp}
y = xW + b,
\end{equation}
where $x \in \mathbb{R}^{(N+1) \times (D_{obs}+D_{agent})}$ is the input, and $W \in \mathbb{R}^{(D_{obs}+D_{agent}) \times D_{out}}$ is the the learnable weights, $b$ is the bias, and $y \in \mathbb{R}^{(N+1) \times D_{out}}$ is the output.
The number of obstacles, $N$, does not affect the parameter of learnable weights.

\subsection{Adaptive Perception Mechanism}\label{sec:APM}
\renewcommand{\algorithmicrequire}{\textbf{Initialization:}}
\renewcommand{\algorithmicensure}{\textbf{Iteration:}}
\newcommand{\Comment}[1]{{\hskip3em$\rightarrow$ #1}}
\begin{algorithm}[htbp]
\caption{AEM Module}
\begin{algorithmic}[1]\label{actalgo}
\REQUIRE ~~\\
Input state = $s_t$, terminal state N{t} = \{$h_{t}^{n}\geq 1-\epsilon$\}, and initialize the hidden state $h_1$ \\

\ENSURE ~~\\
\FOR {n = 1, N}
    \STATE Obtain the interaction features $h_t^n$ with the GRU
    \STATE Compute the believe probability $p_t^n$ of $h_t^n$ according to~\cref{Eq:halt}
    \STATE Update the sum of $p_t^n$ and $p_t^n h_t^n$, respectively
    \IF{{terminal state N(t) or $t \geq t_{max}$}}
        \STATE Break
    \ENDIF
\ENDFOR
 %\STATE Compute the output feature $y_t^n$
\renewcommand{\algorithmicrequire}{\textbf{Output:}}
\REQUIRE ~~\\
 \STATE Return the output feature $y_t^n$ according to~\cref{Eq:aemoutput}
\end{algorithmic}
\end{algorithm}
%\vspace{-0.2cm}
The sigmoidal halting unit, $\sigma(\cdot)$, is added to the end of each GRU, as shown in \cref{fig:rl-model}. It can be used to determine whether the output of the AEM is sufficient to represent the dynamic environment or not through
\begin{equation}\label{Eq:halt}
  p_{t}^{n}=\sigma \left( W^{h}h^{n}_{t}+b^{h} \right),
\end{equation}
where $W^{h}$  represents the weight parameters, $b^{h}$ is the bias parameter {and the parameters of this sigmoidal halting unit are trained within the whole framework}.
The output of the halting unit, given by \cref{Eq:halt}, can also be used as the confidence of environment perception. The adaptive mechanism determines the number of GRUs in use according to the perception confidence and the halt parameter, that is,
\begin{equation}\label{Eq:aemoutput}
%y_{t}^{n} = \frac{1}{N(t)}\sum_{n=1}^{N(t)} p_{t}^{n}h_{t}^{n},
% \begin{array}{c}
\begin{aligned}
y_{t}^{n} &= \frac{1}{N(t)}\sum_{n=1}^{N(t)} p_{t}^{n}h_{t}^{n}, \\
N(t) &= \min\left\{n:\sum_{n=1}^{n} p_{t}^{n}\geq 1-\epsilon\right\},~\textit{s.t.}~n \geq 1,
\end{aligned}
\end{equation}
where $N(t)$ is the adaptive number of GRUs, and $\epsilon$ is the halt threshold, $n$ is the maximum number of GRUs, and the perception confidence is $p_{t}^{n}$. The output of the AEM is $y_t^n \in \mathbb{R}^{(N+1) \times D_{AEM}}$ , given by \cref{Eq:aemoutput}~\cite{ACT2016}. {As we can see, the adaptive perception process is iterative, the final output is the weighted sum-up of all iterations' result}. The overall operation of the AEM module is shown in
\cref{actalgo}.

\subsection{Collision-free Policy Formulation}
\subsubsection{Reward Design}
The reward is expected to be maximally returned by using the optimal policy. Compared to the existing reward functions ~\cite{chen2019crowd,everett2018motion,CADRL2017}, our reward design is different in how to calculate the reward in complex situations and adaptively modulate the reward function according to the dynamic environment. Those reward design methods~\cite{chen2019crowd,everett2018motion,CADRL2017} just consider the closest obstacle and hence are not suitable to deal with variable numbers of obstacles.

\begin{figure}[t]
\centering
   \resizebox{0.9\linewidth}{!}{
   $\begin{array}{cc}
   % Requires
    \subfigure[]{\includegraphics[height=.3\textwidth]{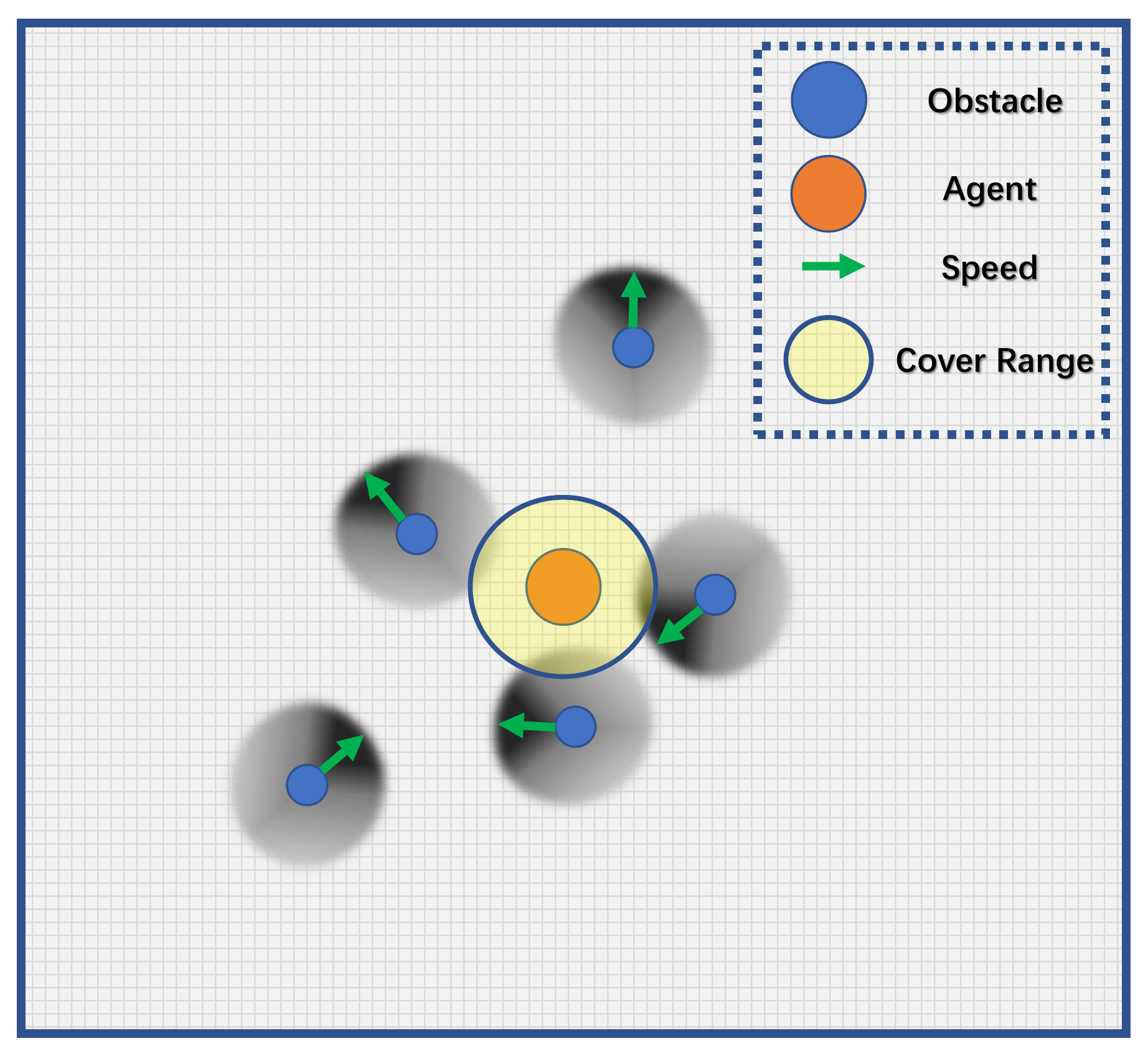} \label{fig.reward_s}} &
   \subfigure[]{\includegraphics[height=.3\textwidth]{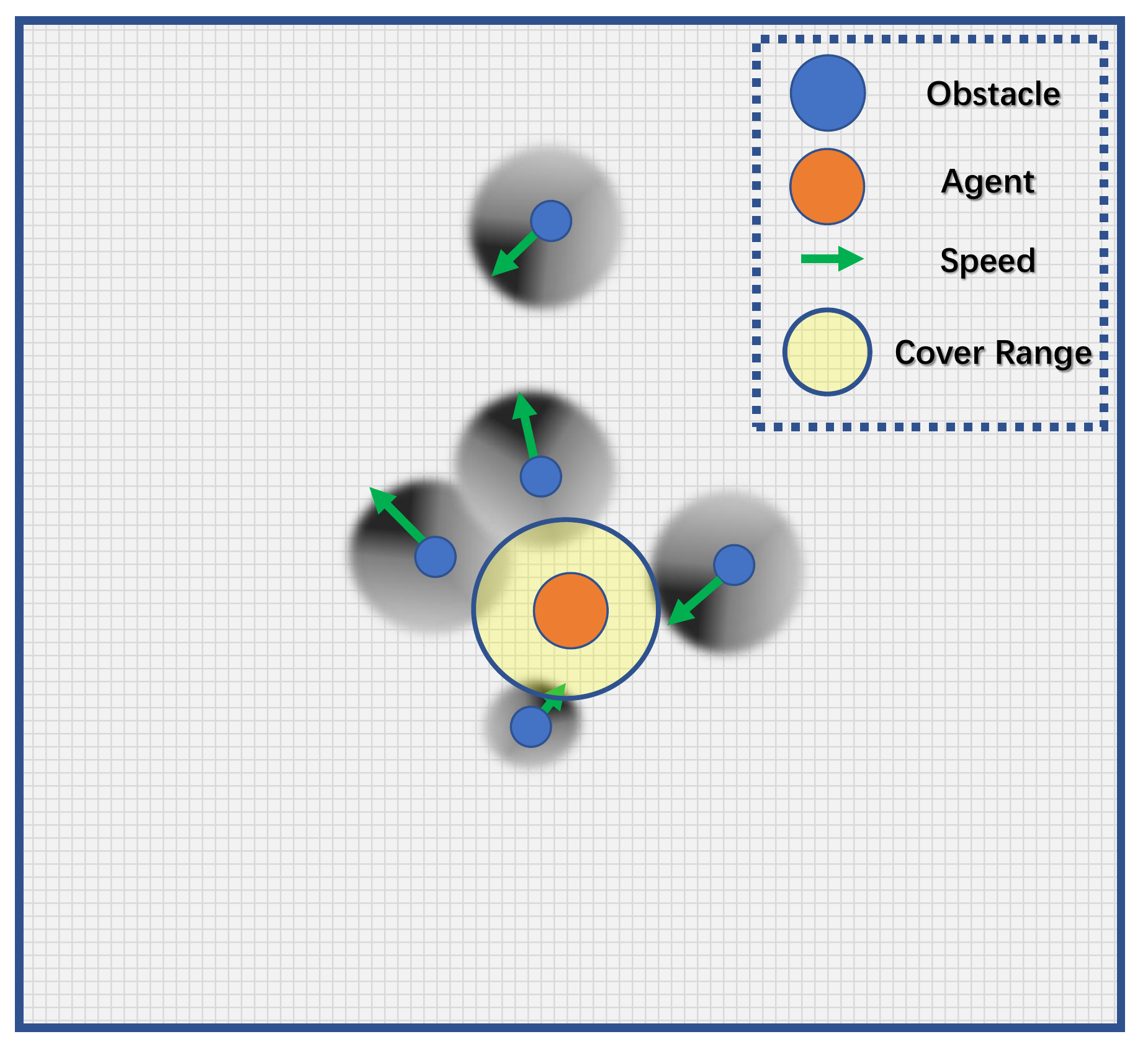} \label{fig.reward_d}} \\
   \end{array}$
   }
  \vspace{-0.3cm}
 \caption{An illustration of the possible collision areas of 5 mobile obstacles moving randomly.
 %The maximum preferred velocity is 1m/s for all and the maximum radius of the obstacles is 0.2m.
  {The shaded areas representing the distribution of obstacles location is proportional to the current speed of each obstacle. The yellow area around the agent represents the agent coverage, therefore the overlap part in the shaded areas and agent coverage is the collision probability}. (a) The obstacles are all in the same velocity with different orientations. The $p_{collision}$ is calculated between the agent and three obstacles. (b) The obstacles are in different velocities and orientations. The $p_{collision}$ is calculated between the agent and four obstacles.}
   \label{fig:prob_reward}
     \vspace{-0.5cm}
    \end{figure}

By contrast, we use the predictive collision possibility ($p_{collision}$) between the agent and obstacles as the reward with unchanged agent and obstacles actions. As shown in~\cref{fig:prob_reward}, we estimate the distribution of obstacles' position in the next time step by current position and velocity, and then the probability of obstacles in the agent's coverage ($\phi_{agent}$) at the next time step is accumulated as the predictive collision possibility ($p_{collision}$). Due to the unpredictable dynamic environment, the reward function cannot be formulated by directly taking the predicted states (const velocity prediction) of the obstacles from the simulator with the long-range time step. And the time step in the reward function is bigger than that of the simulator settings.
% \sout{the predictive collision probabilities are calculated between the agent and three and four obstacles, respectively, which might cause collision in the next step.}
Therefore, our reward function is designed by
\begin{equation}\label{Eq:reward}
R_{t}\left(\mathbf{s}_{t}, \mathbf{a}_{t}\right)=\left\{
\begin{array}{cc}
%{1,} & {\text {if}~\mathbf{p}_{t}=\mathbf{p}_{g}} \\
{1,} & {\text ~\mathbf{Arrival}} \\
{-0.25,} & {\text ~\mathbf{Collision}} \\
% {-0.3 \times (d_s - d_f) / d_s } & {\text { else if } d_{s}<0.2}, \\
{-0.25p_{collision}\beta}, & {\text {\bf{Otherwise}} } \\
%{0} & {\text { otherwise, }}
\end{array}
\right.,
\end{equation}
where $p_{collision} = \sum_{(x,y)\in\phi_{agent}}g_{map}(x,y)$, $\phi_{agent}$ is the coverage of agent as shown in \cref{fig:prob_reward}.
$g_{map}(x,y) $ is the probability that the position $(x,y)$ is occupied by obstacles.
% $\beta$ is the scale ratio of the collision probability within a range from $0.0$ to $2.0$.
$\beta$ is a hyper-parameter that is used to prevent the predictive collision possibility ($p_{collision}$) from being too small to cause the model performance degradation. $\mathbf{Arrival}$ means the distance between the agent and its target position is less than \SI{0.1}{\meter}. At time $t$, $g_{map}(x,y)$ can be computed by
\begin{equation}\label{Eq:grid_map}
\begin{aligned}
 %\vspace{-0.3cm}
    &g_{map}^{t}(x^t,y^t) = \sum_{i=1}^N g^{x}(x^t - x_i^o)
    g^{y}(y^t - y_i^o)g^{\theta}\left(\theta^t - \theta_i^o\right),\\
    &g^x = N(0,\delta_{x}), g^y = N(0,\delta_{y}),g^\theta = N(0,\delta_{\theta}),\\
    & \theta_i^o = \arctan(\frac{v_i^y}{v_i^x}), \theta^t = \arctan(\frac{y^t - y_i^o}{x^t - x_i^o})
    % g_{map}(a^t) = e^{-\frac{(a^t-a_{c}^t)^2}{{\sigma}^2}}
    %  \vspace{-0.1cm}
\end{aligned}
\end{equation}
where N is the number of obstacles, and $\delta_x$, $\delta_y$ and $\delta_\theta$ are hyper-parameters. $(x_i^o,y_i^o)$ is the position of the $i^{th}$ obstacle, and $\theta_i^o$ is the heading angle of the $i^{th}$ obstacle.  $\theta^t$ is the angle between a line from $(x_i^o,y_i^o)$ to $(x^t,y^t)$ and the x-axis. As shown in \cref{Eq:grid_map}, $g_{map}(x,y)$ is the cumulative probability of each obstacle at $(x,y)$.

For each obstacle, the probability that the obstacle is at $(x,y)$ is calculated by ${g^x(x^t-x^o)g^y(y^t-y^o)g^\theta}(\theta^t-\theta^o)$ as shown in \cref{Eq:grid_map}.
To simplify the calculation, we just consider a circular area centered on the current position of the obstacle. The radii of the circular area and $\phi_{agent}$ are $v^o\Delta\mathbf{t_{obstacle}}$ and $v^a\Delta \mathbf{t_{agent}}$ respectively, where $\Delta \mathbf{t_{obstacle}}$ and $\Delta \mathbf{t_{agent}}$ are hyper-parameters which indicate the time prediction length.
%The radius of the circular area is $v^o*\Delta\mathbf{t_{obstacle}}$ where $\Delta \mathbf{t_{obstacle}}$ is a hyper-parameter which indicates the time prediction length.
%And the radius of $\phi_{agent}$ is $v^a * \Delta \mathbf{t_{agent}}$.
As shown in \cref{fig:prob_reward}, circular areas have different radii because obstacles have different speeds. We also build a grid map that the value of each grid is $g_{map}(x,y)$, and this grid map changes over time.

% where $g_{map}(a^t) = e^{-\frac{(a^t-a_{c}^t)^2}{{\sigma}^2}}$, $a$ is a function of $x$, $y$, and $\theta$, and $a_c^t$ is the position of each agent at the time $t$. For simplicity, the variance of position in $\mathbf{X}$ and $\mathbf{Y}$ directions are equal, denoted by $\mathbf{\delta_{xy}}$.
% According to the radius of the possible collision areas, the intersection area between robot and obstacles, can be calculated from all collision coverage areas based grid map.
% It is easy to calculate the total collision grid probability $p_{collision}$ of intersection area between robot and obstacles by the means of grid map.

\subsubsection{Policy Formulation}
\renewcommand{\algorithmicrequire}{\textbf{Initialization:}}
\renewcommand{\algorithmicensure}{\textbf{Iteration:}}
\begin{algorithm}[htbp]
\caption{AEMCARL }
\begin{algorithmic}[1]\label{algo}
\REQUIRE ~~\\
Obtain the terminal state: \{Reaching the goal,  Collision\}\\
Generate training state-action samples $\mathcal{S}$ using the ORCA method, and the memory unit $\mathcal{\hat{M}}\leftarrow\mathcal{S}$ \\
Initialize the value network, $\mathcal{V}\leftarrow\mathcal{S}$, the target value network $\mathcal{\hat{V}} \leftarrow \mathcal{V}$ \\
\ENSURE ~~\\
\FOR {epoch = 1, N }
  \STATE Initialize random samples $s^{0}$ from $\mathcal{\hat{M}}$
  \REPEAT
    \STATE Formulate the interaction feature $s_{IN}$ with the AEM \\
    \STATE Compute the environment feature with the TFM \\
    \STATE Update the value network $\mathcal{V}$ with the action module \\
    \IF{terminal state {\textbf{$s_t$}}}
        \STATE Update the memory unit $\mathcal{\hat{M}}$
    \ENDIF
    \UNTIL{terminal state {\textbf{$s_t$}} or $t \geq t_{max}$}
    \STATE Update the target network $\mathcal{\hat{V}} \leftarrow \mathcal{V}$
\ENDFOR
\renewcommand{\algorithmicrequire}{\textbf{Output:}}
\REQUIRE ~~\\
 \STATE Return $\mathcal{V}$
\end{algorithmic}
\end{algorithm}
% \textcolor{red}{
% The methods~\cite{chen2019crowd,chen2020relational} can successfully deal with the multiple crowded scenarios according to their methods. Both method adopt the attention mechanism based $\mathit{softmax}$ or graph operation. Due to concatenate the agent feature and obstacle feature, the model only obtain the limited interactive information between the agents in the environment.}

%\textcolor{red}{
% The Transformer (TF) Network~\cite{vaswani2017attention} can distill the environment key information with the self-attention mechanism and the multi-head mechanism.

\begin{figure}[t]
 \centering
 %\resizebox{\linewidth}{!}{
 \includegraphics[width=0.4\textwidth]{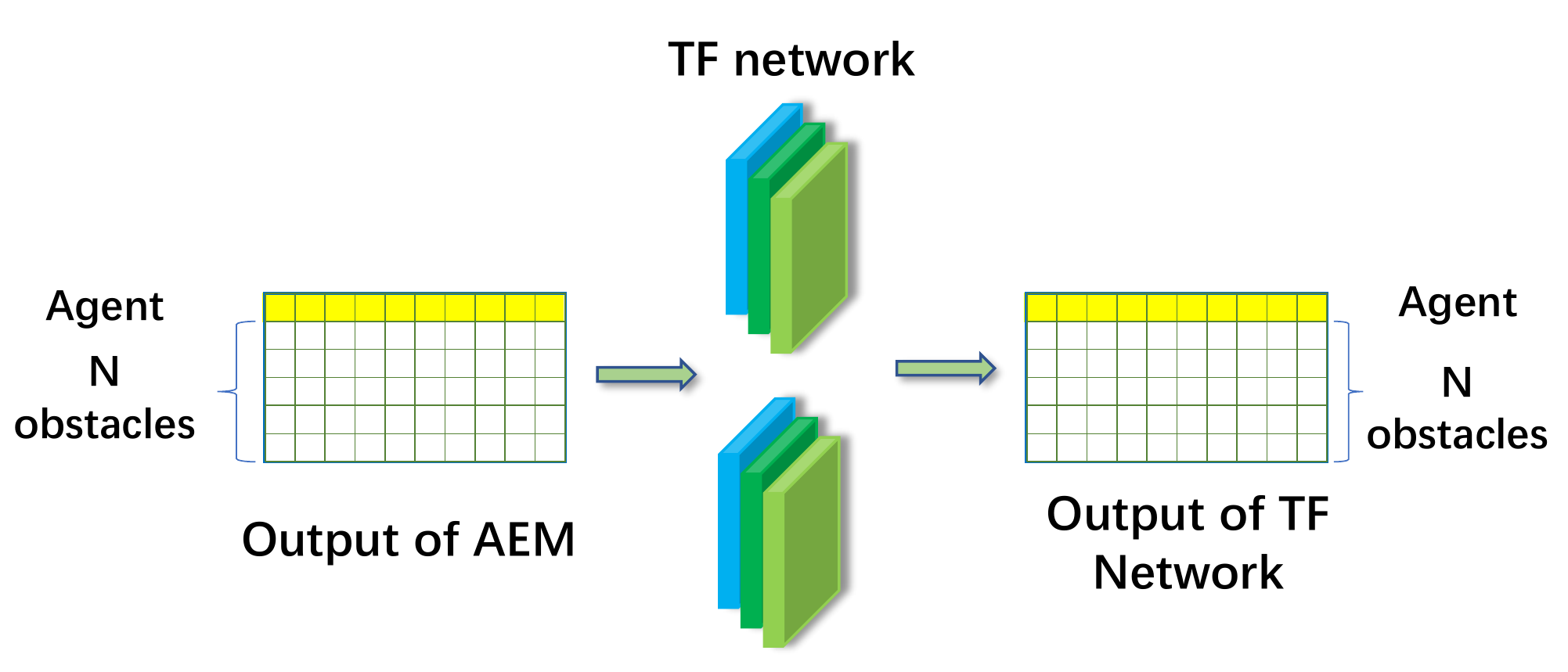}\\ \vspace{-0.3cm}
  \caption{An illustration of the workflow of the TF network. The input of TF network $y_t^n \in \mathbb{R}^{(N+1)*D_{AEM}}$, and TF nwtwork does not change the order of input.
features}
     \label{fig:tfm}
     \vspace{-0.5cm}
\end{figure}

The attention mechanism can help improve the collision-free navigation performance of the model. The Transformer (TF) Network~\cite{vaswani2017attention} can adaptively extract the most important information of features according to the self-attention mechanism, and adopt the multi-head mechanism to estimate more reliable features. The self-attention of the TF module is given by~\cite{vaswani2017attention}
\begin{equation}\label{Eq:TF}
\left\{
\begin{array}{l}
\mathbf{Q}= {\rm MLP}\left( W^q y_t^n + b^q\right), \\
\mathbf{K}= {\rm MLP}\left( W^k y_t^n + b^k\right), \\
\mathbf{V}= {\rm MLP}\left( W^v y_t^n + b^v\right), \\
\mathbf{Att}\left(Q,K,V\right) = {\rm softmax}\left(\frac{QK^T} {\sqrt{d_k}} \right)V ,\\
\end{array}\right.
\end{equation}
where the output order of the AEM modules is from top to bottom. $y_t^n \in \mathbb{R}^{(N+1)*D_{AEM}}$ is the output of AEM. $d_k$ is the dimension of the feature, $Q, K, V$ are three feature embedding layers~\cite{vaswani2017attention}.

As shown in \cref{fig:tfm}, TF network does not change the order of input features and weights each feature with other features by an attention mechanism. Among the $N+1$ features output by TF network, the first feature is the joint state representation of the agent and other obstacles. We only extract the first row of the output of the attention module, $\mathbf{Att}\left(Q,K,V\right)$, as shown the first yellow row at the right of \cref{fig:rl-model} and \cref{fig:tfm}. This is the reason we train in an environment with 5 obstacles but can perform well in an environment with 20 obstacles.
% And the multi-head mechanism can extract the features from different views according to the number of the multi-head, as shown in the Fig.~\ref{fig:rl-model}.}
%This procedure can help increase the speed of decision making and reduce the probability of collision.
%
%An agent sorting block and a Softmax layer are used to construct the attention module (AM). The agent sorting block first assigns weights to obstacles according to their geometric and motion relationships to the robot in the environment. The Softmax layer gives attention scores based on the weights of obstacles. This procedure can help increase the speed of decision making and reduce the probability of collision. The final environment feature weighted by attention scores is denoted by~\textbf{$s_{AM}$}.

%\textcolor{red}{
Therefore, the features of environment, \textbf{$y_t^n$}, are fed into the TF module (TFM) to help achieve collision-free decisions for possible partial observations. The TF output, denoted by $f_{TF}$, is used to represent the feature of multi-agent interactions. The input robot state, $s^t_{agent}$, is concatenated with the embedding feature, $\textit{$f_{TF}$}$, to construct the final environment feature, given by \cref{Eq:actionvalue}.
\begin{equation}\label{Eq:actionvalue}
v^{t}_{action} ={\rm MLP}\left(\left[f_{TF} \oplus s^t_{agent}\right]\right),
\end{equation}
where $\bigoplus$ means the concatenation operation.
The overall procedure of our proposed reinforcement learning of collision-free motion planning policy is shown in \cref{algo}.

\label{trainning}
\subsection{Model Training}
We use the same cost function as DQN to train the framework including AEM (HEM, APM) and TFM. As with other deep learning algorithms, the model is trained by combining the cost function with the gradient backpropagation for parameter updating.
\label{result}
\subsection{Computation Setup}
The key parameters of the RL network include: discount factor $\gamma$ = 0.9, batch size $batch$ = 100, learning rate $L$ = 0.001, halt parameter $\epsilon$=0.05, collision scale ratio $\beta$ = 2.0, and optimism method being Adam~\cite{kingma2014adam}. In our setup, we let $\delta_{x}$ be the same as $\delta_{y}$, denoted as $\delta_{xy}$. The variances of the obstacle position, denoted by $\delta_{xy}$,  and heading angle, denoted by $\delta_\theta$, are set as 2, respectively.
%All the modules in AEMCARL use the multiple EGRUs, multilayer perceptron (MLP) and only one layer of LSTM.
The hidden units of AEM module, TF module, and action module are [(100, 50)], [(150, 150, 150), $nhead$:2], and [(150, 100, 100, 1)], respectively. The action space consists of 80 discrete actions: 5 speeds exponentially distributed over $(0, v^{pref}]$ and 16 orientations evenly distributed over $[0, 2\pi)$.

%As described in the SARL~\cite{chen2018crowd}, CommNet~\cite{CommNet}, the importance of agent depends on the robot-to-agent distance, and velocity \textit{etc.}
%Making a good policy decision depends on understanding both the current scenario and the previous scenario. The information of previous scenarios includes the interaction structure variation and the trajectory of each agent, which contributes to improving the RL model's ability to exploit the past experience. Given the information about the current scenario, the model can give a successful interaction representation based on previous experience. The concatenated features with both importance and interaction relationship attribute are input to the action module, which can contribute to making a successful decision on collision-free policy within the condition of information fully/partially observable. The overall procedure of our proposed reinforcement learning-based collision avoidance method is shown in \textbf{Algorithm~\ref{algo}}.
\section{Experiments}
\label{experiments}
%{\color{red} Specifiy the action types: Discrete, Continuous. Which behavior model is used to generate the movable agents for training? It may significantly influence experimental results. It is unclear that how GSCARL utilized the actor-critic (WRONG definition) framework? Fig. 3 and Alg. 1 only demonstrate the critic network training. Why your approach can improve sample efficiency? The computation process of the network parameters can be provided in supplementary materials. Ablation study is provided to demonstrate the Interaction module.}
The simulation experiments were performed on a PC with an Intel core i7-7700K CPU, an Nvidia GTX1070 GPU, and 32G RAM. In the physical experiments, we used a portable computation platform with an Intel core i5-5500T CPU, and 16G RAM.
There were two testing cases for the unmanned robot navigation with multiple mobile obstacles: $\textit{invisible}$ and $\textit{visible}$. The former means that all obstacles cannot detect the unmanned robot; whereas the latter means that they can.
Since the obstacle robots are operated by the ORCA~\cite{ORCA} method that enables robots to actively avoid obstacles. The invisible case is to ensure that the obstacle robots cannot actively avoid the unmanned robot,  which is used to demonstrate the collision-free performance of the proposed AEMCARL.
\subsection{Quantitative Evaluation}
%Fig.~\ref{fig:traj} illustrates the trajectories of a robot interacting with 5 mobile obstacles in the invisible case. Fig.~\ref{Fig.g-circle} shows a simulation experiment with the Gym-Gazebo platform with 12 mobile obstacles.
 %That requires the collision-free method should instantly make a successful decision to keep the robot in safe.
%Fig.~\ref{fig:traj} shows the trajectories of a robot interacting with the other movable agents in an invisible test case.
\begin{table}
  %\vspace{-0.1cm}
  \centering
  \caption{A COMPARISON OF NAVIGATION PERFORMANCE BETWEEN THE STATE-OF-THE-ART METHODS AND OUR METHOD}
  \resizebox{\linewidth}{!}{
  \begin{tabular}{c|c|c|c|c}
     \hline  \hline
     Methods & Success Rate& Collision Rate& Time (s) & Timeout Rate\\ \hline
     ORCA~\cite{ORCA} & 0.43 & 0.57 & 10.93 & $\textbf{0.00}$ \\ \hline
     LM-SARL-Linear~\cite{chen2019crowd} & 0.90 & 0.09  & 11.15 & 0.01 \\ \hline
     RGL-Linear~\cite{chen2020relational} & 0.91 & 0.01 & \textbf{10.37} & 0.08 \\ \hline
     MP-RGL-Onestep~\cite{chen2020relational} & 0.92 & 0.02  & 10.44 & 0.06\\ \hline
     MP-RGL-MultiStep~\cite{chen2020relational} & 0.97 & 0.01  & $10.95$ & 0.02 \\ \hline
     %LM-SARL & $\textbf{1.00 / 1.00}$ & $\textbf{0.00 / 0.00}$ & 10.46 / 10.59  \\ \hline
     AEMCARL (Ours) & $\textbf{1.00}$  & $\textbf{0.00}$ & 11.09 & $\textbf{0.00}$ \\
     \hline \hline
  \end{tabular}
  }
  \begin{tabular}{p{8cm}}
    All experiments are based on one invisible unmanned robot and five mobile obstacles. The AEMCARL in table adopts the designed adaptive reward function.
    %The experiments used one invisible unmanned robot and 5 mobile obstacles. All experiments used one invisible unmanned robot and 5 mobile obstacles.
  \end{tabular}
  \label{tab:invisible-result}
  \vspace{-0.3cm}
\end{table}

% \begin{table}
%   %\vspace{-0.1cm}
%   \centering
%   \caption{A COMPARISON OF NAVIGATION PERFORMANCE BETWEEN THE STATE-OF-THE-ART METHODS AND OUR METHOD}
%   \resizebox{\linewidth}{!}{
%   \begin{tabular}{c|c|c|c|c}
%      \hline  \hline
%      % after \\: \hline or \cline{col1-col2} \cline{col3-col4} ...
%      Methods & Success Rate & Collision Rate& Time (s) & Timeout Rate\\ \hline
%      ORCA~\cite{ORCA} & 43.0\% & 57.0\% & 10.93 & $\textbf{0.0\%}$ \\ \hline
%      LM-SARL-Linear~\cite{chen2019crowd} & 90.0\% & 9.0\%  & 11.15 & 1.0\% \\ \hline
%      RGL-Linear~\cite{chen2020relational} & 91.0\% & 1.0\% & \textbf{10.37} & 8.0\% \\ \hline
%      MP-RGL-Onestep~\cite{chen2020relational} & 92.0\% & 2.0\%  & 10.44 & 6.0\%\\ \hline
%      MP-RGL-MultiStep~\cite{chen2020relational} & 97.0\% & 1.0\%  & $10.95$ & 2.0\% \\ \hline
%      %LM-SARL & $\textbf{1.00 / 1.00}$ & $\textbf{0.00 / 0.00}$ & 10.46 / 10.59  \\ \hline
%      AEMCARL (Ours) & $\textbf{100.0\%}$  & $\textbf{0.0\%}$ & 11.09 & $\textbf{0.0\%}$ \\
%      \hline \hline
%   \end{tabular}
%   }
%   \begin{tabular}{p{8cm}}
%     All experiments are based on one invisible unmanned robot and 5 mobile obstacles. The AEMCARL in table adopts the designed adaptive reward function.
%     %The experiments used one invisible unmanned robot and 5 mobile obstacles. All experiments used one invisible unmanned robot and 5 mobile obstacles.
%   \end{tabular}
%   \label{tab:invisible-result}
%   \vspace{-0.2cm}
% \end{table}

The invisible case is more challenging for collision-free navigation than that of the visible case.
\cref{tab:invisible-result} shows a comparison between the state-of-the-art methods and our method. To validate the performance of the proposed method, the baseline methods are selected by following the rule that all state-of-the-art (SOTA) methods are trained by the same simulator and trained from scratch by ourselves. The comparison is based on 5 obstacles for invisible cases with 500 random experiments. It can be seen that our method is the most robust one to reach the goal with the smallest collision probability. The last column of \cref{tab:invisible-result} shows that our method can accomplish the task each time within the predefined time which is set as 20 seconds.
As shown in~\ref{Fig.sr}, the RGL~\cite{chen2020relational} methods are with the poorest generalization for the scenes with more obstacles (e.g. $n \geq 5$ ). As shown in~\cref{tab:invisible-result}, the proposed method outperforms the other baseline methods with narrow improvements and total success rate. However, the proposed method can also keep the robustness to the more complex scenes that include more obstacles of different sizes shapes, and velocities, as shown in~\cref{Fig.sr}.

The $\textbf{Time}$ metric in~\cref{tab:invisible-result} means the average time with 500 random experiments. Except for the time limits of RGL variants methods being set 30 seconds, the other methods are 20 seconds. The $\textbf{Time-out}$ metric means the ratio of the experiments without reaching the destination in the allowed maximum time of the methods, accordingly.
%During in the multi-agents scenarios, our proposed method, named AEMCARL, outperforms the ORCA, CADRL, and LSTM-RL.

\begin{figure}[t]
% \vspace{-0.2cm}
\resizebox{\linewidth}{!}{
$\begin{array}{cc}
% Requires
%\subfigure[]{\includegraphics[height=.25\textwidth]{human5-10-15-20_gru_rate_inv} \label{Fig.rate_inv}} &
%\subfigure[]{\includegraphics[height=.25\textwidth]{human5-10-15-20_gru_rate_visible} \label{Fig.rate_v}} &
\subfigure[]{\includegraphics[height=.25\textwidth]{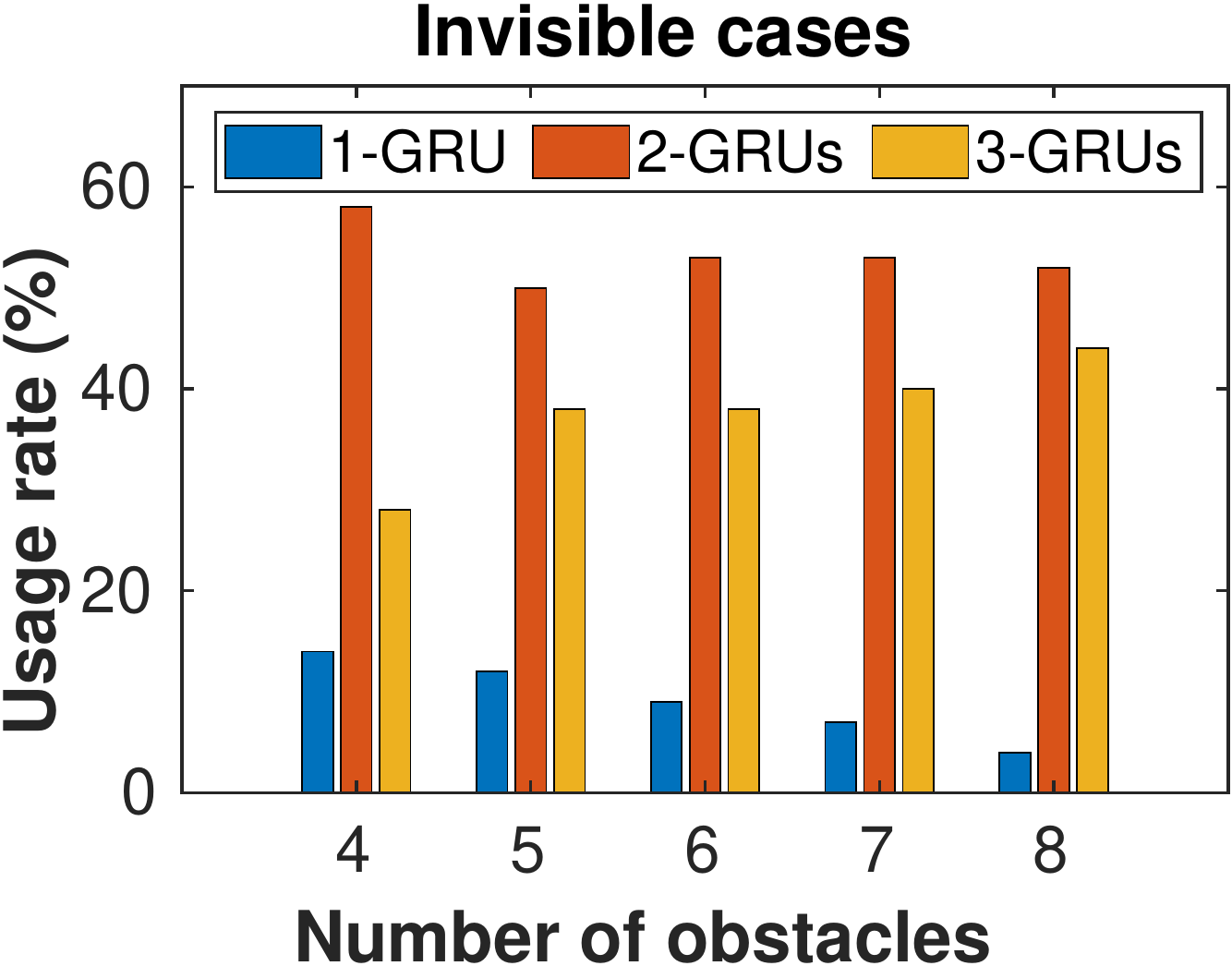} \label{Fig.rate_inv}} &
\subfigure[]{\includegraphics[height=.25\textwidth]{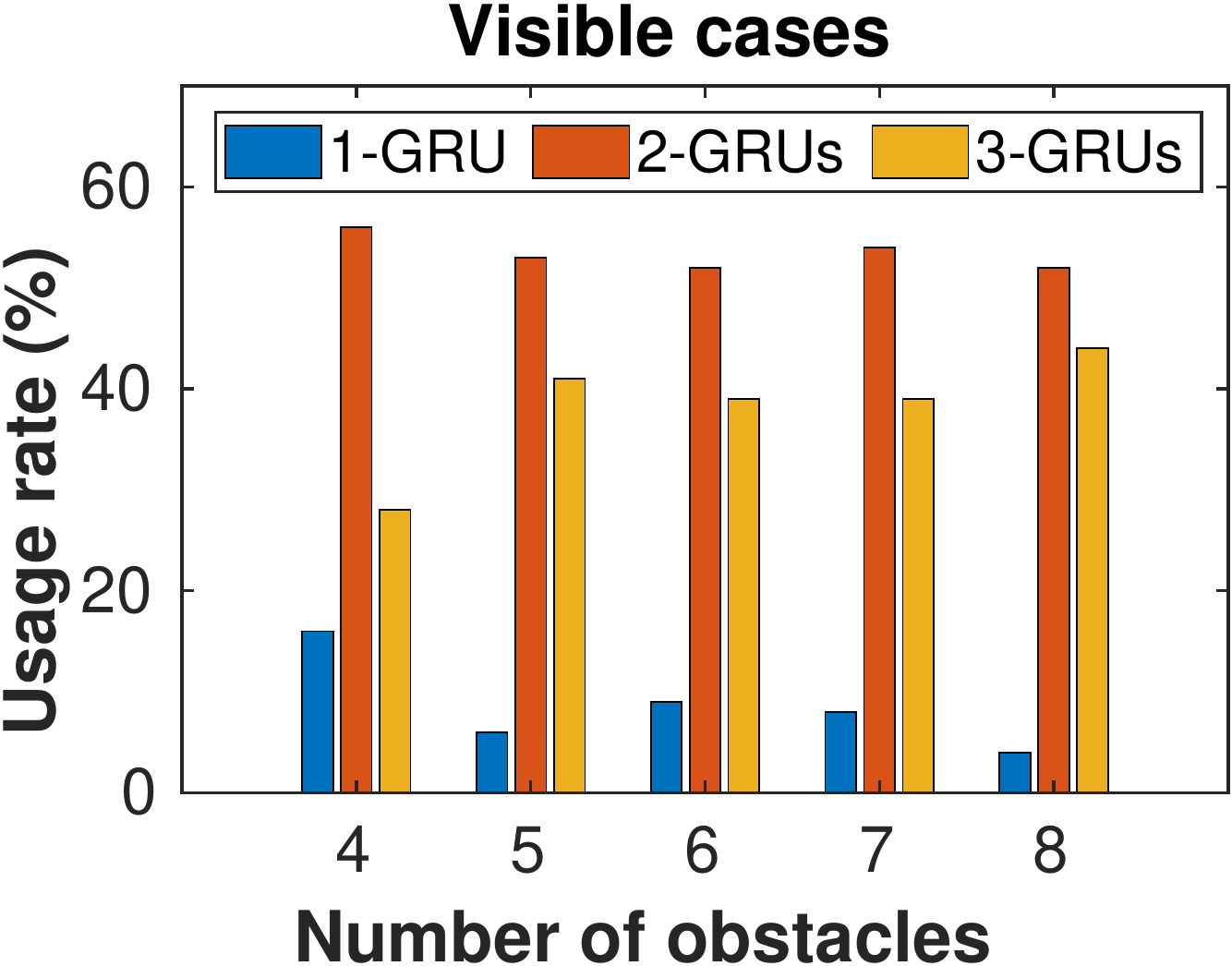} \label{Fig.rate_v}} \\
\subfigure[]{\includegraphics[height=.25\textwidth]{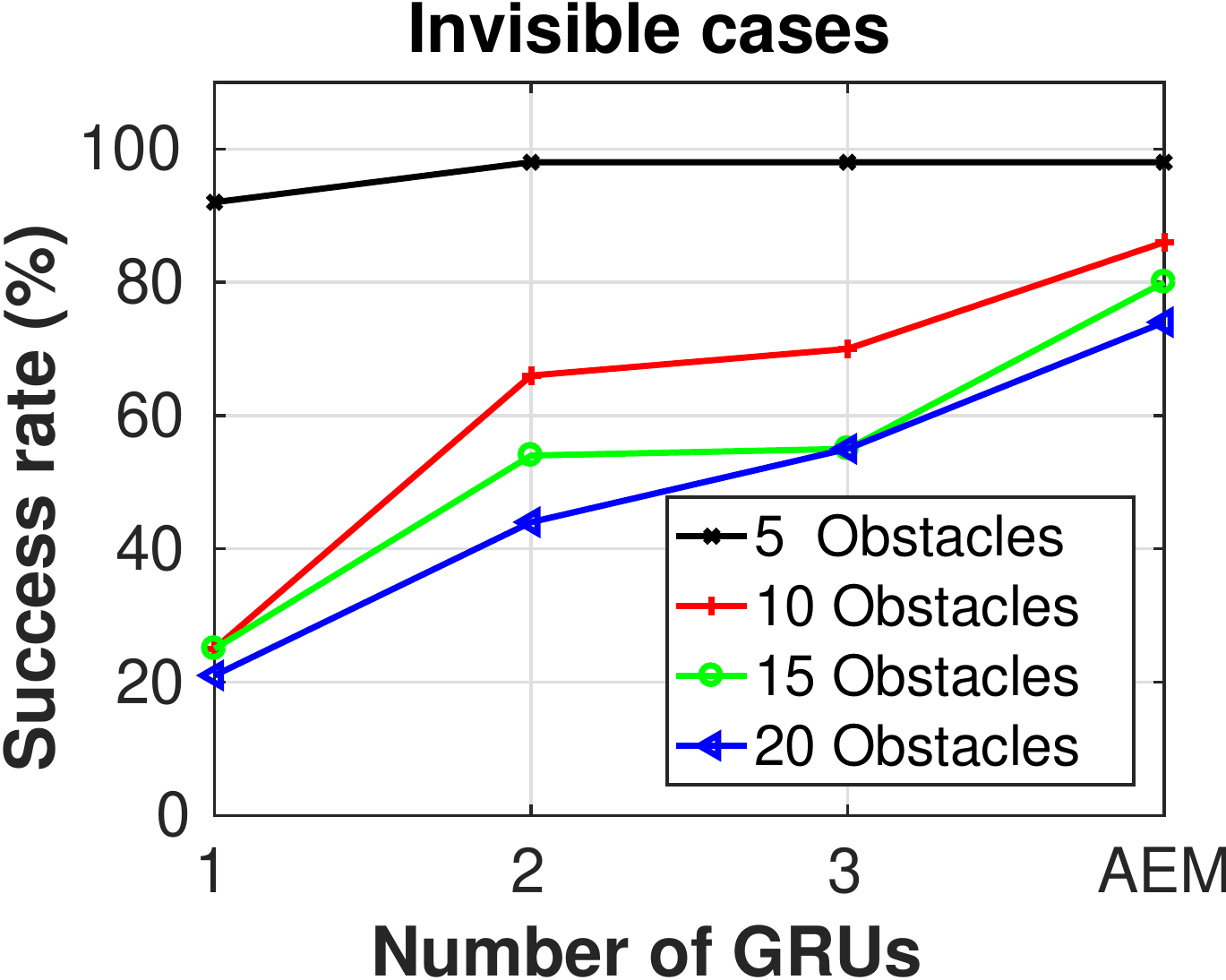} \label{Fig.sr-ivcase}} &
\subfigure[]{\includegraphics[height=.25\textwidth]{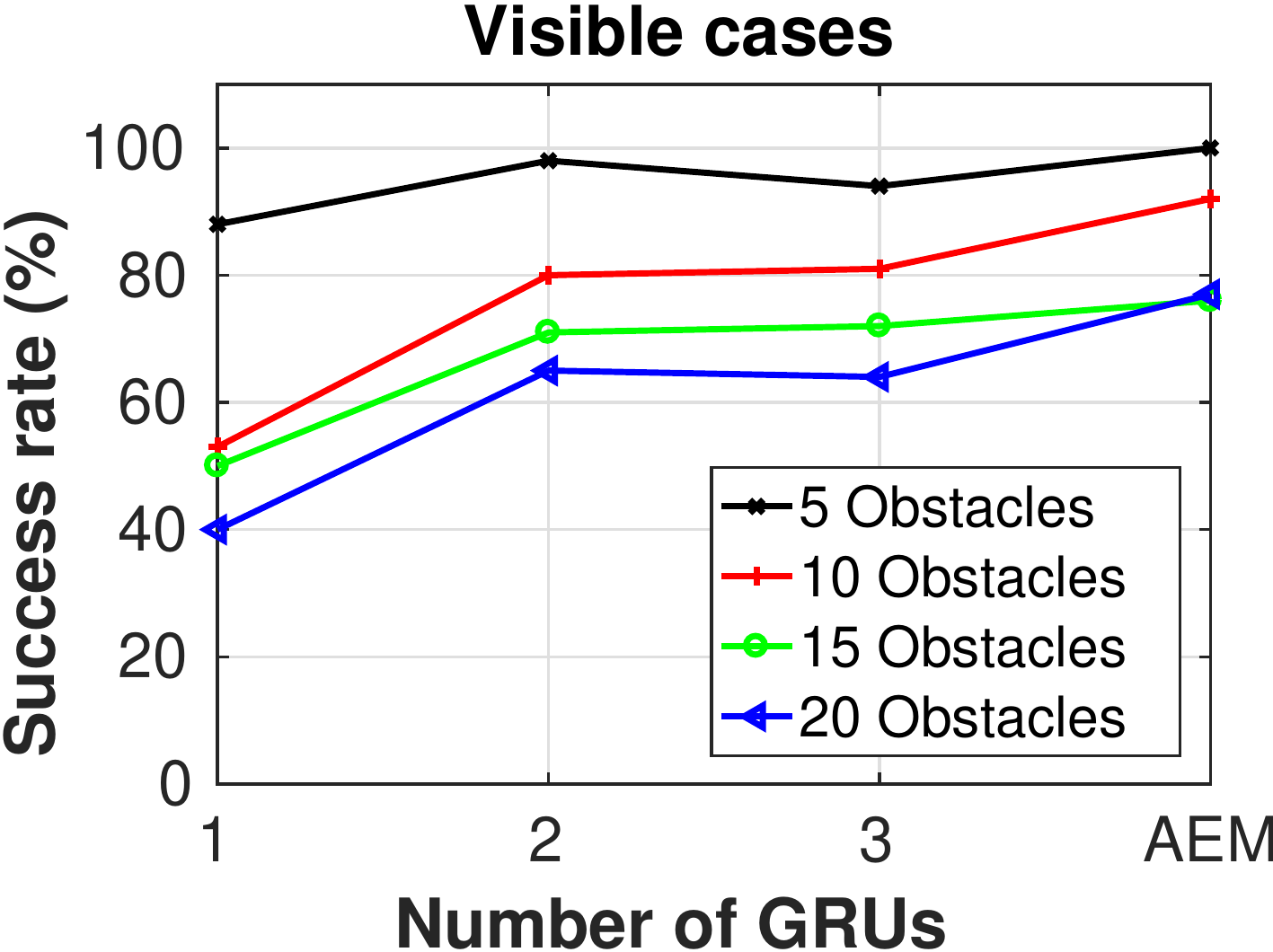} \label{Fig.sr-vcase}}\\
\end{array}$
}
\vspace{-0.2cm}
\caption{An illustration of AEM operations in AEMCARL. (a \& b) An illustration of the usage rates of GRUs in AEMCARL for
different numbers of obstacles in invisible (a) and visible (b) cases, respectively. (c \& d) The success rates of the AEMCARL using different fixed numbers of GRUs and the adaptive mechanism (AEM), for different numbers of obstacles in invisible (c) and visible (d) cases, respectively.}
\label{fig:rate}
\vspace{-0.5cm}
\end{figure}

\subsection{Ablation Study}
\subsubsection{AEM Efficiency}
% \begin{figure*}[htpb]
% % \vspace{-0.2cm}
% \resizebox{\linewidth}{!}{
% $\begin{array}{cccc}
% % Requires
% %\subfigure[]{\includegraphics[height=.25\textwidth]{human5-10-15-20_gru_rate_inv} \label{Fig.rate_inv}} &
% %\subfigure[]{\includegraphics[height=.25\textwidth]{human5-10-15-20_gru_rate_visible} \label{Fig.rate_v}} &
% \subfigure[]{\includegraphics[height=.25\textwidth]{egru-rate_inv_4_8} \label{Fig.rate_inv}} &
% \subfigure[]{\includegraphics[height=.25\textwidth]{egru-rate_v_4_8} \label{Fig.rate_v}} &
% \subfigure[]{\includegraphics[height=.25\textwidth]{sr-cases-iv-gru} \label{Fig.sr-ivcase}} &
% \subfigure[]{\includegraphics[height=.25\textwidth]{sr-cases-v-gru} \label{Fig.sr-vcase}}\\
% \end{array}$
% }
% \vspace{-0.4cm}
% \caption{An illustration of AEM operations in AEMCAR. (a \& b) An illustration of the usage rates of GRUs in AEMCARL for
% different numbers of obstacles in invisible (a) and visible (b) cases, respectively. (c \& d) The success rates of the AEMCARL using different fixed numbers of GRUs and the adaptive mechanism (AEM), for different numbers of obstacles in invisible (c) and visible (d) cases, respectively.}
% \label{fig:rate}
% \vspace{-0.2cm}
% \end{figure*}

 \begin{figure*}[htpb]
%  \vspace{-0.1cm}
   \resizebox{\textwidth}{!}{
   $\begin{array}{ccc}
   % Requires
      \subfigure[]{\includegraphics[height=.3\textwidth]{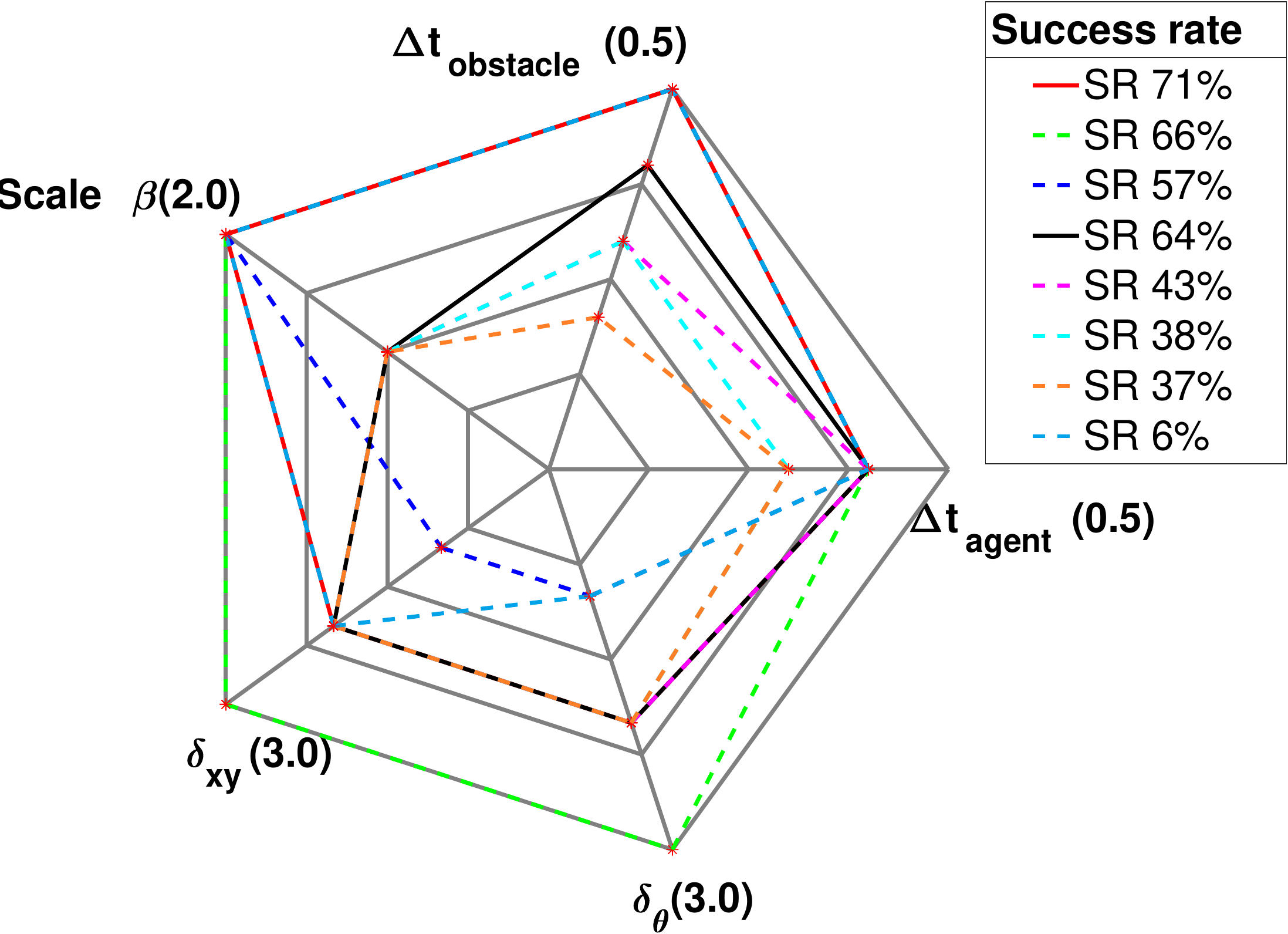} \label{Fig.reward_paras}} &
    \subfigure[]{\includegraphics[height=.3\textwidth]{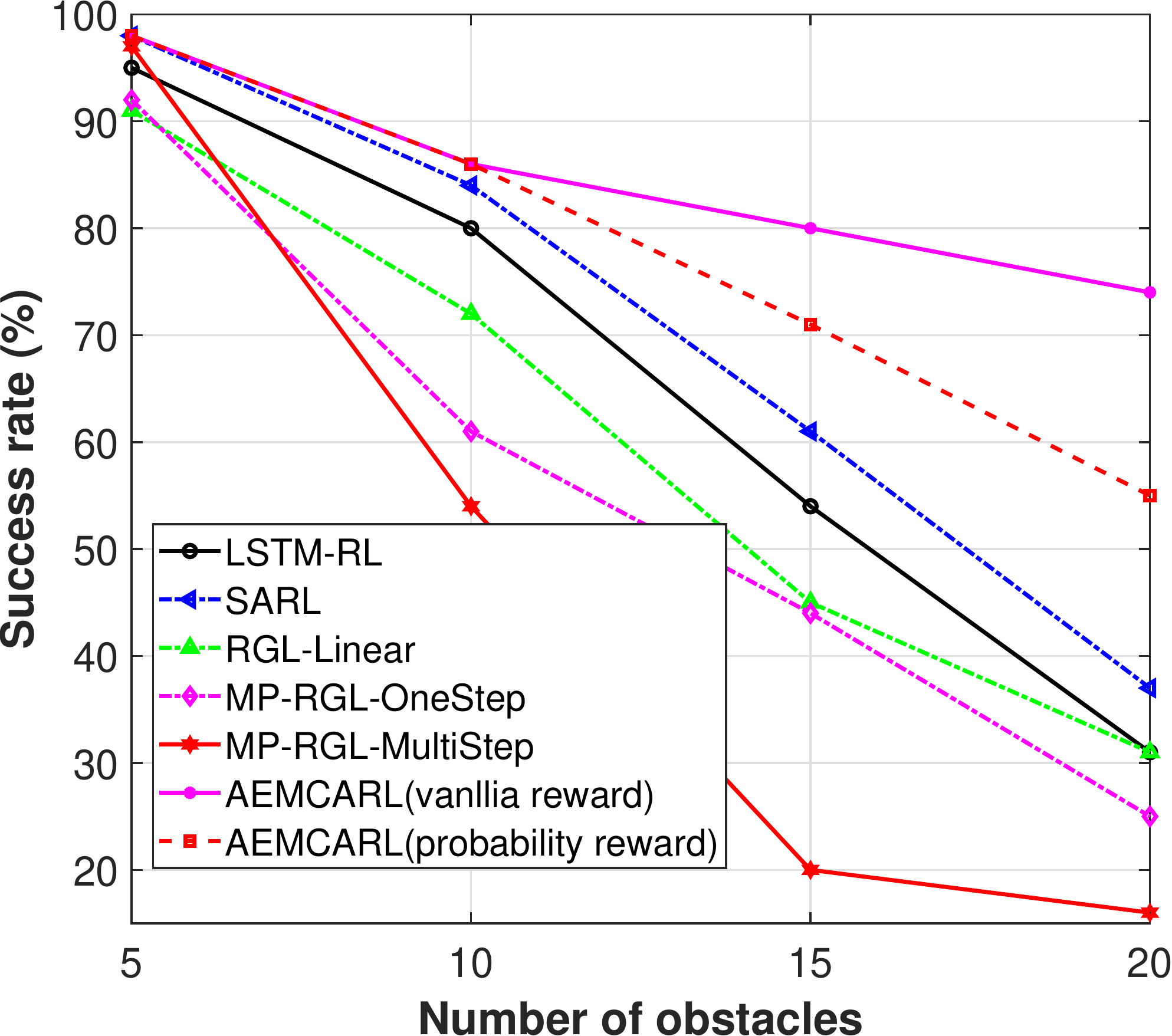} \label{Fig.sr}} &
   \subfigure[]{\includegraphics[height=.3\textwidth]{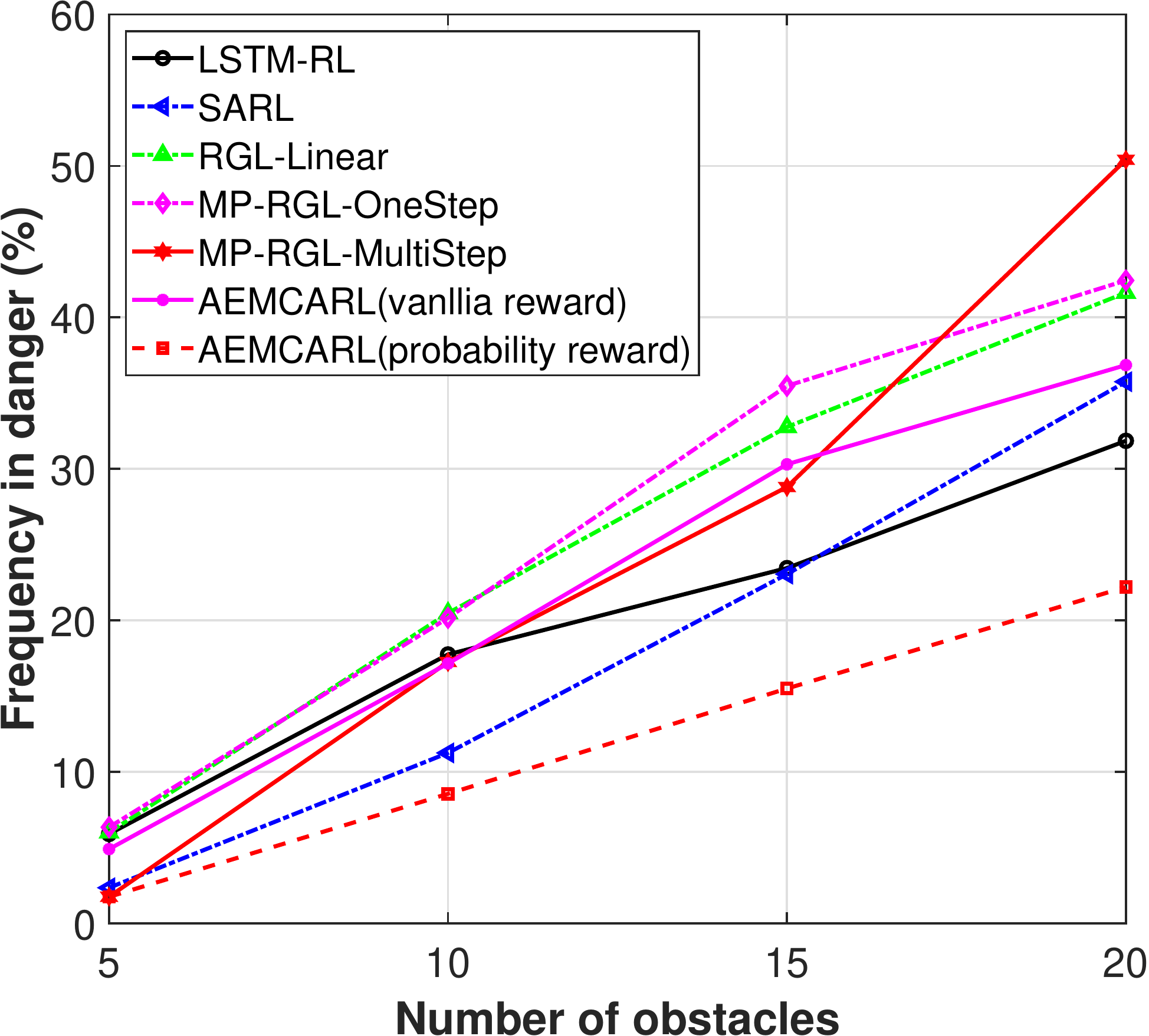} \label{Fig.dist}} \\
   \end{array}$
   }
   \vspace{-0.5cm}
 \caption{The evaluation of AEMCARL. (a) A comparison of the navigation success rates using different parameters in the case of 20 obstacles with 50  random experiments. The number in the bracket means the maximum value of the corresponding radius. (b) A comparison of the navigation success rates using LSTM-RL, SARL, RGL-Linear, MP-RGL-OneStep, MP-RGL-MultiStep, and AEMCARL for different numbers of obstacles. (c) A comparison of frequency in danger using LSTM-RL, SARL, RGL-Linear, MP-RGL-OneStep, MP-RGL-MultiStep, and AEMCARL for different numbers of obstacles.}
   \label{fig:exp-table}
  \vspace{-0.5cm}
    \end{figure*}
Our approach was also tested with real-world experiments in different scenarios using 4 to 8 mobile obstacles, respectively. Turtlebots were used as mobile obstacles and a Husky was used as the unmanned robot that performed the RL-based collision-free navigation, as shown in \cref{fig:poster}.
% All robots are equipped with UWB devices for %distance measurements and localization.
\cref{Fig.rate_inv} and \cref{Fig.rate_v} show the usage rate of 3 GRUs for different scenarios in the real world, which can be seen in the appendix video material\footnote{ https://github.com/SJWang2015/AEMCARL}. It can be seen that the proposed adaptive perception module (APM) can automatically change the structure of the hierarchical environment module (HEM) in different scenarios to optimize the framework of the neural network.

Both SARL and AEMCARL use the attention mechanism to focus on the most threatening obstacles, which helps them to achieve fast training convergence and best navigation performance in the case of a different number of obstacles (\cref{tab:invisible-result}).
%  By comparison, AEMCARL has fewer parameters (90\% of SARL), and hence runs faster.
 The adaptive perception module can further help AEMCARL to improve the computation speed without losing the navigation performance.
 \cref{Fig.sr-ivcase} and \cref{Fig.sr-vcase} show a comparison of navigation performance between using 3 different fixed numbers of GRUs and the adaptive perception mechanism (AEM) given a different number of obstacles for 100 random experiments of both visible and invisible cases. It can be seen that the AEM outperforms any fixed number of GRUs.
 %For dynamic environment representation, SARL only uses s part of the local information and lacks a full understanding of the overall environment, which may decrease the model's ability to exploiting known information.
%{\color{red}ADD AN TABLE FOR DIFFERENT STEP DEMENSTRATION}

%\subsubsection{Experiment results.}
%The difference between AEMCARL and other's method is that AEMCARL can deal with more complicated scenarios. With the increase number of agents, AEMCARL smartly control the Husky robot with a slow speed to avoid the collision. In previous works, the robot almost automatically runs in a environment with low-speed or small range of speed agents, like a corridor with a walking person, \textit{etc.} However, the robot in this work is automatically performed in a scenario which includes the walking pedestrians, and high-speed running cars.
%A hardware video is available at XXXXXX.
% \addtolength{\textheight}{-12cm}
\subsubsection{Reward Design}
% \begin{figure}[htpb]
%   \vspace{-0.2cm}
%   \resizebox{\linewidth}{!}{
%   $\begin{array}{cc}
%   % Requires
%     \subfigure[]{\includegraphics[height=.2\textwidth]{invisible-multihuman-sr_5_10_15_20} \label{Fig.sr-ivcase}} &
%   \subfigure[]{\includegraphics[height=.2\textwidth]{visible-multihuman-sr_5_10_15_20} \label{Fig.sr-vcase}}\\
%   \end{array}$
%   }
%  \caption{The success rates of the AEMCARL using different fixed numbers of GRUs and the adaptive mechanism for 5, 10, 15, and 20 obstacles in invisible and visible cases, respectively.}
%   \label{fig:ablation}
%     \end{figure}

%The reward function plays an important role in the RL methods. It can decide the performance of the RL methods.
The proposed adaptive reward function contains multiple parameters, as shown in \cref{Fig.reward_paras}. To study the role of each parameter in the reward function, we conducted a set of experiments to compare the performance of each parameter with 100 random experiments under the model trained by only 2000 episodes.
The success rate was used as the performance indicator. \cref{Fig.reward_paras} shows that the parameters, including $\Delta$$\mathbf{t_{agent}}$, $\Delta$$\mathbf{t_{obstacle}}$, and $\mathbf{Scale~\beta}$, basically are proportional to the success rate. Since the hyperparameters ($\mathbf{\delta_{xy}}$, $\mathbf{\delta_{\theta}}$) are used to tune the expectation of the squared deviation of the mean of the velocity and heading angle. So the relationships between $\mathbf{\delta_{xy}}$$\setminus$$\mathbf{\delta_{\theta}}$ and the success rate are not monotonous.

\subsubsection{Policy Efficiency}
Both \cref{Fig.sr} and \cref{Fig.dist} show the comparison of system performance for different numbers of obstacles in the same size between three open-source baseline RL methods (LSTM-RL, SARL and RGL) and our method with 100 random experiments.
~\cref{Fig.sr} shows the success rate of baseline methods dramatically degrades with the increase of the number of obstacles.
The vanilla reward in \cref{Fig.sr} is the same as SARL. Compared to the vanilla reward, our adaptive reward based probability with the highest success rate in the cases of a different number of obstacles.
In addition, the AEMCARL with the vanilla reward outperforms the state-of-the-art (SOTA) methods, SARL and RGL, which can demonstrate the efficiency of the framework of our model. SARL and RGL use two different methods to represent the obstacles states in the environment: the attentive pooling mechanism and the relational graph learning approach, respectively. Our proposed reward function and model framework can adapt to different environments, which can increase the robustness of the collision-free RL model.
\subsubsection{Policy Robustness}
The proposed method, AEMCARL, was trained in the gym simulator with 5 obstacles, but our trained policy, AEMCARL, was tested in the environment cases with multiple obstacles $\left(\geq 5\right)$. As shown in \cref{tab:invisible-result}, \cref{Fig.sr} and \cref{Fig.dist}, our proposed method outperforms the SOTA methods in 5, 10, 15, and 20 obstacles environments without tuning, which can demonstrate the robustness of the proposed method.

The frequency in danger means the ratio between the number of minimum agent-to-object separation distances lower than the threshold and the total number of actions. It can reflect the degree of aggressiveness policies when processing the danger cases.
Thanks to AEM, adaptive reward function, and TF module, \cref{Fig.sr} and \cref{Fig.dist} show that our method outperforms the two baseline methods in terms of both success rate and frequency in danger.
\section{CONCLUSION}
\label{conclusions}
This paper has presented an adaptive environment modeling-based reinforcement learning (AEMCARL) network which focuses on (1) dynamic environment representation, (2) adaptive perception mechanism, and (3) adaptive reward function for collision-free motion planning policy. The proposed hierarchical environment model (HEM) can robustly and
efficiently model the dynamic environment.
The adaptive perception mechanism (APM) can adaptively use computational resources within a certain degree of perception confidence. The reward function can adaptively modulate the reward
according to the relation between the robot with the real-time environment.  The simulation experiment results show that by training with 5 obstacles and testing with 20 ones, (1) our proposed algorithm can
achieve at least 74\% of success rate, which is 37\% higher than the SOTA algorithms, and achieve the lowest frequency in danger; (2) compared with policies using a different fixed number of GRUs, our AEM can get the best performance in terms of success rate. Our future work will further investigate
the computational efficiency as well as relationships between perception/planning complexities of the environment and RL models.

%have shown that our method can achieve collision avoidance up to 99$\%$ success rate within less than 10 agents, 97$\%$ success rate with 15 agents, and 84$\%$ success rate with 20 agents in test case,
%\section*{ACKNOWLEDGMENT}
%This work is partially supported by the National Natural Science Foundation of China (No: 61773197), the Science and Technology Innovation Committee of Shenzhen City (No: GJHZ20170314114424152), , the Nanshan District Science and Technology Innovation Bureau (No: LHTD20170007).
%%%%%%%%%%%%%%%%%%%%%%%%%%%%%%%%%%%%%%%%%%%%%%%%%%%%%%%%%%%%%%%%%%%%%%%%%%%%%%%%%
%%\begin{thebibliography}{1}
%\newpage

\bibliographystyle{IEEEtran}
\bibliography{MyReference}

%%\end{thebibliography}

\end{document}